\documentclass[10pt,twocolumn,letterpaper]{article}

\usepackage{wacv}              %

\usepackage{float}
\usepackage{graphicx}
\usepackage{graphicx}
\usepackage{amsmath}
\usepackage{relsize}
\usepackage{amssymb}
\usepackage{booktabs}
\usepackage{color}
\usepackage{multirow}
\usepackage[T1]{fontenc}
\usepackage{bm} 
\usepackage{emo}
\usepackage{subcaption}
\usepackage[percent]{overpic}
\usepackage[accsupp]{axessibility}

\usepackage{ulem}
\usepackage{xcolor}
\usepackage{todonotes}

\newcommand{\kk}[1]{{\color{black}#1}}
\newcommand{\change}[1]{\textcolor{black}{#1}}

\newcommand{\loss}[1]{\ell_\text{#1}}

\newcommand{\radiance}{\mathbf{c}}
\newcommand{\shadowedradiance}{\tilde{\radiance}}
\newcommand{\albedo}{\bm{\rho}}
\newcommand{\gaussians}{\mathcal{G}}
\newcommand{\latents}{\mathcal{E}}
\newcommand{\splatting}{\mathcal{S}}
\newcommand{\normal}{\mathbf{n}}
\newcommand{\embedding}{\mathbf{e}}
\newcommand{\sceneindex}{c}
\newcommand{\rotation}{\mathbf{R}}
\newcommand{\translation}{\mathbf{t}}
\newcommand{\scaling}{\mathbf{s}}
\newcommand{\opacity}{o}
\newcommand{\light}{L}
\newcommand{\lightsh}{\mathbf{l}}
\newcommand{\transferfun}{D}
\newcommand{\transferfunsh}{\mathbf{d}}
\newcommand{\params}{\bm{\theta}}
\newcommand{\lossweight}[1]{\lambda_{\text{#1}}}
\newcommand{\image}{\mathcal{I}}
\newcommand{\camera}{\mathcal{C}}
\newcommand{\direction}{\bm{\omega}}
\newcommand{\bx}{\mathbf{x}}

\newcommand{\supplementary}{\texttt{Supplementary}}
\DeclareMathOperator*{\argmin}{arg\,min}

\usepackage[pagebackref,breaklinks,colorlinks]{hyperref}

\newcommand{\ours}{LumiGauss}

\usepackage[capitalize]{cleveref}
\crefname{section}{Sec.}{Secs.}
\Crefname{section}{Section}{Sections}
\Crefname{table}{Table}{Tables}
\crefname{table}{Tab.}{Tabs.}

\def\wacvPaperID{347} %
\def\confName{WACV}
\def\confYear{2025}

\newcommand{\inputimage}{Undefined}
\newcommand{\inputimageshadows}{Undefined}

\def\customparskip{.5em}
\renewcommand{\paragraph}[1]{\vspace{\customparskip}\noindent\textbf{#1}.}

\usepackage{tikz}
\usetikzlibrary{calendar,fpu,matrix, positioning,arrows,arrows.meta,calc,decorations.pathreplacing,decorations.text,spy}
\tikzset{
  img/.style={
    inner sep=0pt,     %
    outer sep=0pt,     %
    rectangle,
    align=center} %
}

\tikzset{
    moon colour/.style={
        moon fill/.style={
            fill=#1
        },
        scale=0.5,
    },
    sky colour/.style={
        sky draw/.style={
            draw=#1
        },
        sky fill/.style={
            fill=#1
        }
    },
    southern hemisphere/.style={
        rotate=180
    }
}

\makeatletter
\pgfcalendardatetojulian{2010-01-15}{\c@pgf@counta} %
\def\synodicmonth{29.530588853}
\newcommand{\moon}[2][]{%
    \edef\checkfordate{\noexpand\in@{-}{#2}}%
    \checkfordate%
    \ifin@%
        \pgfcalendardatetojulian{#2}{\c@pgf@countb}%
        \pgfkeys{/pgf/fpu=true,/pgf/fpu/output format=fixed}%
        \pgfmathsetmacro\dayssincenewmoon{\the\c@pgf@countb-\the\c@pgf@counta-(7/24+11/(24*60))}%
        \pgfmathsetmacro\lunarage{mod(\dayssincenewmoon,\synodicmonth)}
        \pgfkeys{/pgf/fpu=false}%
    \else%
        \def\lunarage{#2}%
    \fi%
    \pgfmathsetmacro\leftside{ifthenelse(\lunarage<=\synodicmonth/2,cos(360*(\lunarage/\synodicmonth)),1)}%
    \pgfmathsetmacro\rightside{ifthenelse(\lunarage<=\synodicmonth/2,-1,-cos(360*(\lunarage/\synodicmonth))}%
    \tikz [moon colour=white,sky colour=black,#1]{
        \draw [moon fill, sky draw] (0,0) circle [radius=1ex];
        \draw [sky draw, sky fill] (0,1ex)
            arc (90:-90:\rightside ex and 1ex)
            arc (-90:90:\leftside ex and 1ex)
            -- cycle;
    }%
}

\newcommand{\shadowed}{\moon{8}}
\newcommand{\unshadowed}{\moon{16}}

\makeatletter
\apptocmd\@maketitle{{\teaserfigure{}\par}}{}

\makeatother

\newcommand{\teasermanual}{
\begin{subfigure}[b]{\linewidth}
    \centering
    \resizebox{\linewidth}{!}{
        \begin{tikzpicture}[
            >=stealth',
            overlay/.style={
              anchor=south west, 
              draw=black,
              rectangle, 
              line width=0.8pt,
              outer sep=0,
              inner sep=0,
            },
            font=\Huge
        ]
        \matrix[matrix of nodes, column sep=1pt, row sep=6em, ampersand replacement=\&, inner sep=0, outer sep=0, font=\Huge] (training) {
            \includegraphics[width=\linewidth, trim={0 50 0 225}, clip]{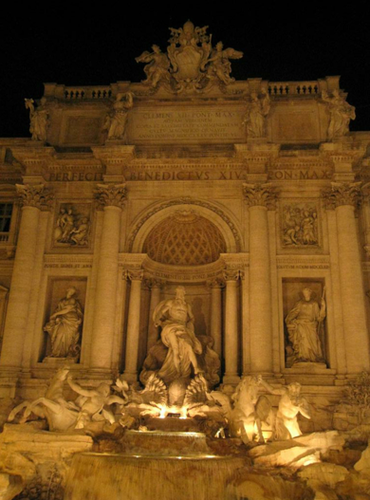} \&
            \includegraphics[width=\linewidth, trim={0 50 0 225}, clip]{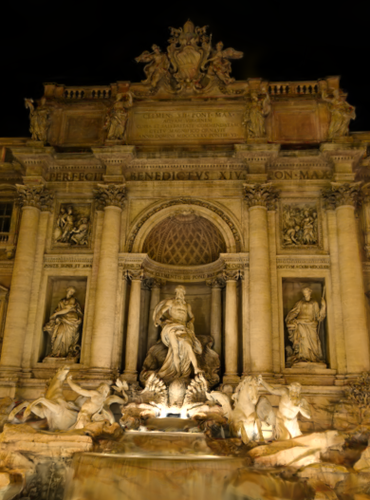} \&
            \includegraphics[width=\linewidth, trim={0 50 0 225}, clip]{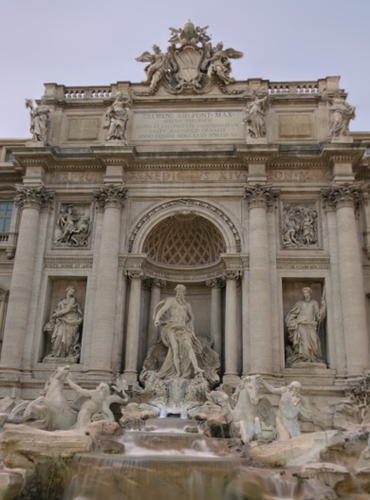} \&
            \includegraphics[width=\linewidth, trim={0 50 0 225}, clip]{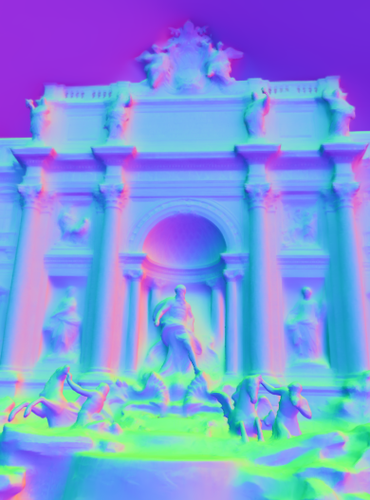} \\
            \includegraphics[width=\linewidth]{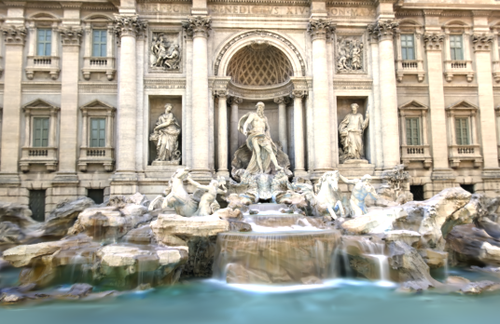} \&
            \includegraphics[width=\linewidth]{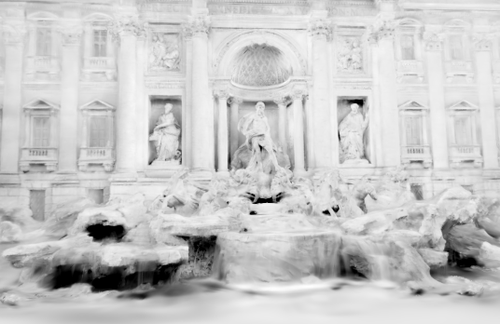} \&
            \includegraphics[width=\linewidth]{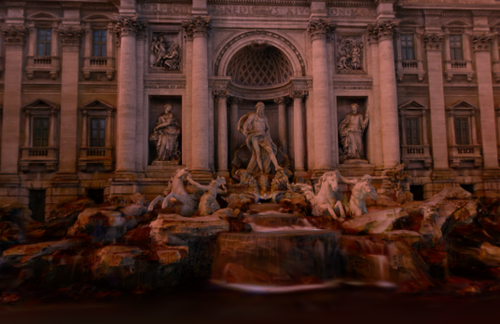} \&
            \includegraphics[width=\linewidth]{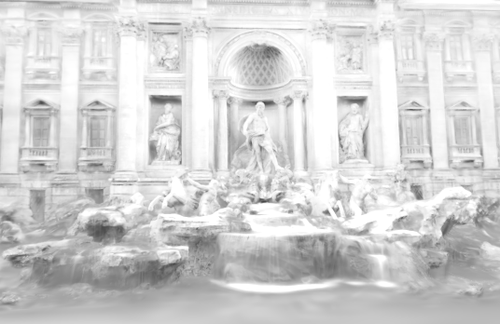} \\
        }; 
        
        \node[above=1.6em of training-1-1.north, anchor=base,scale=1.3] {Training Image};
        \node[above=1.6em of training-1-2.north, anchor=base,scale=1.3] {Reconstruction};
        \node[above=1.6em of training-1-3.north, anchor=base,scale=1.3] {Predicted Albedo};
        \node[above=1.6em of training-1-4.north, anchor=base,scale=1.3] {\change{Rendered} Normals};
        
        \node[above=1.6em of training-2-1.north, anchor=base,scale=1.3] {Novel Light};
        \node[above=1.6em of training-2-2.north, anchor=base,scale=1.3] {Predicted Approximate Shadows};
        \node[above=1.6em of training-2-3.north, anchor=base,scale=1.3] {Novel Light};
        \node[above=1.6em of training-2-4.north, anchor=base,scale=1.3] {Predicted Approximate Shadows};

        \node[below right=5em and 1pt of training-2-1.north east, anchor=center]{\includegraphics[width=4em]{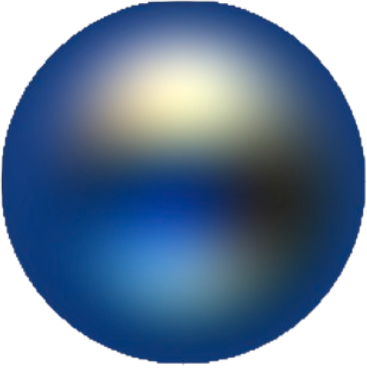}};
        \node[below right=5em and 1pt of training-2-3.north east, anchor=center]{\includegraphics[width=4em]{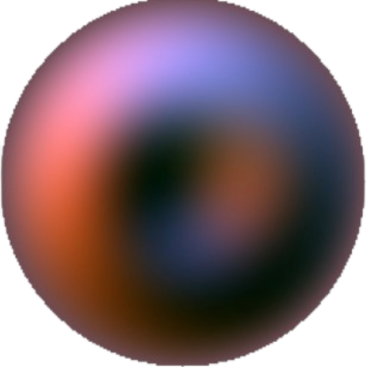}};
    \end{tikzpicture}
    }
\end{subfigure}
}
\newcommand{\teaserfigure}{
\captionsetup{type=figure}

\fboxsep=0pt %
\fboxrule=0.4pt %

\vspace{-1.5em}
\teasermanual
\vspace{-1.7em}

\setcounter{figure}{0} %
\captionof{figure}{
{%
\textbf{Teaser} --
\ours{} reconstructs environment maps and object surfaces from \textit{in-the-wild} images. Our model decouples the scene color and its normals (\textit{second and fourth column in the top row}). At inference, it can synthesize novel views (\textit{bottom row}) and realistic lighting (\textit{first and third columns in the bottom}) with high-fidelity shadows (\textit{second and fourth columns in the bottom}).
}
}
\vspace{1em}
\label{fig:teaser}
}

\begin{document}

\title{\ours: Relightable Gaussian Splatting in the Wild}

\author{
 \textbf{Joanna Kaleta}$^{1,2}$\thanks{Corresponding authors: joanna.kaleta.dokt@pw.edu.pl} %
 \quad
 \textbf{Kacper Kania}$^{1}$%
 \quad
 \textbf{Tomasz Trzci{\'n}ski}$^{1,4,5}$
 \quad
 \textbf{Marek Kowalski$^{3}$} \\
$^1$\small Warsaw University of Technology \quad $^2$\small Sano Centre for Computational Medicine \quad
$^3$\small Microsoft \quad 
$^4$\small IDEAS NCBR \quad 
$^5$\small Tooploox \quad
}

\maketitle

\begin{abstract}
Decoupling lighting from geometry using unconstrained photo collections is notoriously challenging. Solving it would benefit many users as creating complex 3D assets takes days of manual labor. Many previous works have attempted to address this issue, often at the expense of output fidelity, which questions the practicality of such methods. We introduce \ours{} - a technique that tackles 3D reconstruction of scenes and environmental lighting through 2D Gaussian Splatting. Our approach yields high-quality scene reconstructions and enables realistic lighting synthesis under novel environment maps. We also propose a method for enhancing  the quality of shadows, common in outdoor scenes, by exploiting spherical harmonics properties. Our approach facilitates seamless integration with game engines and enables the use of fast precomputed radiance transfer. We validate our method on the NeRF-OSR dataset, demonstrating superior performance over baseline methods. Moreover, \ours{} can synthesize realistic images for unseen environment maps. Our code: \url{https://github.com/joaxkal/lumigauss}.
\end{abstract}

\section{Introduction}
\label{sec:introduction}

The colors emitted by objects are a combination of a spectrum of the light hitting the object and the material properties of that object.
The light hitting the object's surface is a sum of the light scattered in the medium and bounced from neighboring objects \cite{whitted1979improved}.
In computer graphics, we often simplify this effect and decouple it into two entities: an intrinsic object's color or \textit{albedo} and an omnidirectional texture representing the illumination \cite{ramamoorthi2001envmap}---\textit{environment map}. Acquiring those assets enables the designing of realistic scenes in games or movies.

In many scenarios, creating realistic albedo textures and environment maps requires skilled technicians and artists to be involved in the process. To democratize it, the previous approaches~\cite{rudnev2022nerfosr, gardner2023neusky, wang2023fegr} tried to use photographs taken with commodity cameras and \textit{invert} the capturing process to recover albedo and an environment map. 
Given the abundance of casual, in-the-wild photographs available on the Internet, solving that issue is of high importance.

Recent advancements in reconstruction in-the-wild include NeRF-in-the-Wild~\cite{martin2021nerfw} (NeRF-W). NeRF-W leverages neural radiance fields~\cite{mildenhall2021nerf} which reconstruct a scene given its photos with calibrated cameras. NeRF-W can further work in realistic scenarios where the pictures come from the \textit{in-the-wild} collections---the images in such may differ in the lighting conditions or scene content. However, NeRF-W and its follow-up works, HA-NeRF~\cite{chen2022hallucinated} and CR-NeRF~\cite{yang2023crnerf}, cannot decouple the object's albedo and the environment map, making it difficult to use in practice.  NeRF-OSR~\cite{rudnev2022nerfosr} approaches that problem, but its shading model requires neural network execution at runtime, making integration with graphics engines difficult, and the reconstruction quality leaves space for improvement.

3D Gaussian Splatting~\cite{kerbl20233dgaussiansplatting} (3DGS) solves one of the main bottlenecks of NeRF - the training speed and output fidelity. In contrast to NeRFs, 3DGS models the scene as a composition of 3D Gaussians attributed with colors and opacity which are rasterized, or \textit{splatted}, to render the output image. Recovering an object's surface from them requires specialized training techniques~\cite{guedon2024sugar}. On the other hand, 2DGS~\cite{huang20242dgs} proposes reformulating 3D Gaussians as their 2D alternative where one of the axes is collapsed. The final scene representation ends up being composed of 2D \textit{surfels} which provide a flat surface crucial for our relighting approach.

In this work, we propose \ours{}, a method that uses  2DGS~\cite{huang20242dgs} to perform inverse graphics on images taken in the wild.
In contrast to past approaches, our method is imbued with fast training and inference speed while maintaining high-quality renderings and being easy to integrate with graphics engines. %
In our method, the light is modeled as a combination of an environment map and a radiance transfer function that represents which parts of the environment map illuminate a given surfel---both are modeled by spherical harmonics~\cite{ramamoorthi2001envmap}. This approach allows for modeling shadows, which is our main goal, but also has the potential to represent light reflected off of other objects. 
The output from \ours{} enables both novel view synthesis and relighting using environment maps beyond those available during training. Leveraging the possibilities offered by the precomputed radiance transfer, our representation integrates seamlessly into game engines, enabling fast and efficient relighting.

Our contributions:
\begin{itemize}
    \item We repurpose 2D Gaussian Splatting for an inverse graphics pipeline in an in-the-wild setting. With our approach, we recover high-quality albedo and environment maps.
    \item To enable shadows we learn the radiance transfer function for each 2D splat and represent it using spherical harmonics.
    \item Finally, we demonstrate that our reconstructed environment maps can be effectively used to relight arbitrary objects within graphic engines.
\end{itemize}

\section{Related Works}
\label{sec:related_works}

\paragraph{Relighting} Relighting outdoor scenes is a key challenge in computer graphics and VR/AR. Early works \cite{lalonde2009webcam, troccoli2005relighting, haber2009relighting, xing2013lighting, sunkavalli2007factored, duchene2015multi, barron2014shape} used training-free methods like statistical inference. Deep learning approaches, such as Yu~\etal~\cite{yu2020self} with a neural renderer, and Philip~\etal~\cite{philip2019multi} with proxy geometry, face limitations in reconstruction quality and viewpoint flexibility.

NeRF-based methods~\cite{mildenhall2021nerf} enabled simultaneous viewpoint and lighting changes. However, methods like~\cite{zhang2021nerfactor, zeng2023nrhints, srinivasan2021nerv} handle a single illumination only or specific illumination setup during. Others are object-specific, such as for faces \cite{sun2021nelf}. Many unconstrained photo collection methods focus on appearance, not lighting, complicating integration with other graphical components~\cite{martin2021nerfw, chen2022hallucinated, yang2023crnerf, li2023msnerf}. 

NeRF-based approaches, such as \cite{pun2023lightsim} and \cite{urbanir}, focus on inverse rendering for outdoor scenes, particularly in applications like autonomous driving. However, these methods are designed for single video sequences rather than unstructured photo collections. Rudnev~\etal~\cite{rudnev2022nerfosr} proposed a method for relighting landmarks from unconstrained photo collections, using NeRF with external lighting extraction. Similarly, \cite{li2022neulighting} compresses the per-image illumination into a disentangled latent vector. Wang~\etal~\cite{wang2023fegr} target static scenes and works with unconstrained photo collections but rely on costly mesh extraction. Some methods incorporate additional priors, environmental assumptions, or regularizations~\cite{solnerf, yang2023complementary}. Gardner~\etal~\cite{gardner2023neusky} leverage externally trained models to provide environmental lighting priors. Despite their potential, these methods cannot be used in real-time applications due to NeRF’s slow training and rendering times.

\change{In contrast, the TensoRF-based approach by Chang~\etal~\cite{chang2024srtensorf} aligns time information and sun direction with images for relighting, eliminating the need for external lighting models. While this method is faster than NeRF, it still lacks seamless integration with graphics engines and is unsuitable for synthetic light integration.}

Notable Gaussian Splatting works designed for unconstrained photo collections \cite{kerbl20233dgaussiansplatting, dahmani2024swagsplattingwildimages, zhang2024gaussian, xu2024wildgs, wang2024wegs} focused on appearance editing, not seamless graphical component integration. Relightable Gaussian approaches, like  \cite{gao2023relightable, liang2024gs, shi2023gir}, tackle material decomposition but are not adapted to handle varying lighting conditions of \textit{in-the-wild} training setup. Radiance transfer properties, employed in a similar way to \ours{}, are utilized in \cite{zhang2024prtgaussianefficientrelightingusing, saito2024relightablecodecavatar}. However, these methods rely on a burdensome dataset setup, restricting their applicability to specific use cases.

\paragraph{Gaussian Splatting} Kerbl~\etal~\cite{kerbl20233dgaussiansplatting} introduced a notion of using learnable 3D Gaussian primitives from point clouds. Those Gaussians are parametrized with 3D covariance matrix $\Sigma_k$ and their location $\translation_k$:
\begin{equation}
    \gaussians(\translation) = \exp(\frac{1}{2}(\translation - \translation_k)^\top\Sigma_k^{-1}(\translation - \translation_k)),
\end{equation}
where the covariance matrix is factorized into a scaling diagonal matrix $\scaling_k$ and a rotation matrix $\rotation_k$ as $\Sigma_k{=}\rotation_k\scaling_k\scaling_k^\top\rotation_k^\top$. An image is rendered with a splatting operator $\splatting(\cdot)$ which projects Gaussians into the camera coordinates with a world-to-camera matrix and then to image plane with a local affine transformation~\cite{zwicker2001ewa}:
\begin{equation}
\label{eq:splatting-op}
    \splatting(\camera_c\;|\;\gaussians) = \sum_{k=1}^K\radiance_k\opacity_k\gaussians_k\prod^{k-1}_{j=1}(1 - \opacity_j \gaussians_k).
\end{equation}
The operator produces an RGB image, given a calibrated camera matrix $\camera_c$ and their additional Gaussians' attributes: their colors $\radiance$ and opacities $\opacity$. Attributes are learned using a stochastic gradient descent.

Huan~\etal~\cite{huang20242dgs} argues that 3DGS although producing high-quality images, the implicit surface representation is noisy, limiting its applicability in relighting scenarios. They propose using 2D Gaussians instead to create smooth, coherent meshes thanks to their exact 2D surfel projection. \kk{We leverage that representation in our \ours{}---a relightable model that decouples albedo, environment light and shadows thanks to our proposed physical constraints}.

\definecolor{fixedop}{HTML}{6E8DBC}
\definecolor{learnableop}{HTML}{82B26C}
\newcommand{\myscheme}{
\centering
\resizebox{\linewidth}{!}{
    \begin{tikzpicture}[
        >=stealth',
        shorten >= 2pt,
        shorten <= 2pt,
        thick,
        align=center,
        fixed/.style={
            rectangle,
            draw=fixedop,
            thick,
            line width=0.5mm,
            rounded corners=4pt
        },
        learnable/.style={
            rectangle,
            draw=learnableop,
            thick,
            line width=0.5mm,
            rounded corners=4pt
        },
    ]

    \node (inputimages) {\includegraphics[width=4cm]{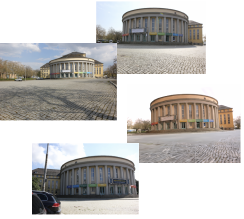}};
    \node[fixed, above=0.5cm of inputimages.north, text width=4cm] (cameras) {calibrated\\cameras from SfM\\$\{\camera_c\}$};
    \node[right=1cm of cameras.east]  (rawgausses) {\includegraphics[trim={2px 0 0 0}, clip]{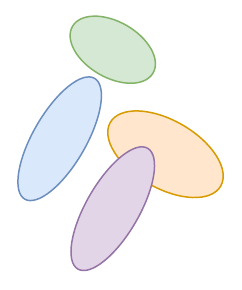}};
    \node[learnable,above right=1cm and 1cm of rawgausses.east, minimum width=2cm] (normal) {normal $\normal_k$};
    \node[learnable,above right=0cm and 1cm of rawgausses.east, minimum width=2cm] (albedo) {albedo $\albedo_k$};
    \node[learnable,above right=-1cm and 1cm of rawgausses.east, minimum width=2cm] (sh) {SH $\transferfunsh_k$};

    \node[fixed, above right=-1cm and 4cm of rawgausses.east, minimum width=3cm, minimum height=1cm] (shadowed) {shadowed\\(\cref{eq:shadowed-radiance})}; 
    \node[fixed, above right=0.5cm and 4cm of rawgausses.east, minimum width=3cm, minimum height=1cm] (unshadowed) {unshadowed\\(\cref{eq:unshadowed-radiance})}; 

    \node[right=8cm of rawgausses.east] (coloredgausses) {\includegraphics{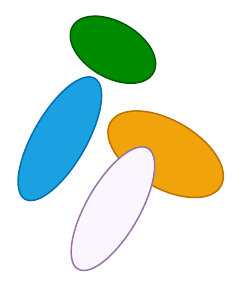}};
    \node[below=0.5cm of coloredgausses.south] (rendered) {\includegraphics{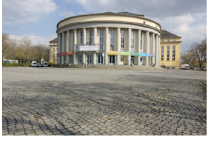}};

    \node[below=0em of inputimages.south, text width=5cm, anchor=north] (unconstrained) {Unconstrained photo collection\\$\{\image_c\}$};

    \node[learnable, text width=4cm, right=1cm of inputimages] (envmap) {Environment map embedding \\(\cref{eq:lightsh})};
    \node[right=2cm of envmap.east] (shball) {\includegraphics[trim={2pt 2pt 2pt 2pt}, clip]{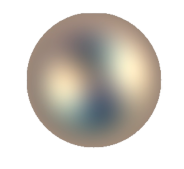}};

    \node[below=-0.5em of rendered.south, text width=4cm,anchor=north] {Render image\\with novel view and light};

    \node[fixed, right=2cm of unconstrained.east, minimum height=0.7cm, minimum width=4cm] (fixedopnode) {fixed operation};
    \node[learnable, right=0cm of fixedopnode.east, minimum height=0.7cm, minimum width=4cm] (learnableopnode) {learnable operation};

    \draw [->] (inputimages.north) -- (cameras.south);
    \draw [->] (cameras.east) -- (rawgausses.west);
    \draw [->] (coloredgausses.south) to  node[right] {$\splatting(\cdot)$}  (rendered.north);
    \draw [->, rounded corners=2pt] (rawgausses.east) |- (normal.west);
    \draw [->, rounded corners=2pt] (rawgausses.east) |- (albedo.west);
    \draw [->, rounded corners=2pt] (rawgausses.east) |- (sh.west);

    \draw [->, rounded corners=2pt] (normal.east) -- ($(normal.east)+(12pt, 0)$) |-  (unshadowed.west);
    \draw [->, rounded corners=2pt] (albedo.east) -- ($(albedo.east)+(12pt, 0)$) |- (unshadowed.west);
    \draw [->, rounded corners=2pt] (albedo.east) --  ($(albedo.east)+(12pt, 0)$) |- (shadowed.west);
    \draw [->, rounded corners=2pt] (sh.east) --  ($(sh.east)+(12pt, 0)$) |-  (shadowed.west);

    \draw [->, rounded corners=2pt] (unshadowed.east) -- ($(unshadowed.east)+(12pt, 0)$) |- (coloredgausses.west);
    \draw [->, rounded corners=2pt] (shadowed.east)-- ($(shadowed.east)+(12pt, 0)$) |- (coloredgausses.west);

    \draw [->] (inputimages.east) -> (envmap.west);
    \draw [->] (envmap.east) -> (shball.west);
    
    \draw [->, rounded corners=2pt] (shball.north) |- ([yshift=0.2cm]shadowed.south west);
    \draw [->, rounded corners=2pt] (shball.north) |- ([yshift=0.2cm]unshadowed.south west);

    \node[above=-1em of rawgausses.north] {{Raw 2D Gaussians} \cite{zhang2024gaussian}};
    \node[above=-1em of coloredgausses.north] {{Relighted 2D Gaussians} (\textbf{Ours)}};
\end{tikzpicture}
}
}

\begin{figure*}[!t]
    \centering
    \myscheme{}
    \vspace{-2em}
    \caption{%
        \textbf{Pipeline --}
        \ours{} learns the relightable 2D Gaussian~\cite{zhang2024gaussian} representation from unconstrained photo collection with variable camera parameters and lighting conditions. Each of $k$ Gaussians holds: a normal $\normal_k$, albedo $\albedo_k$, and learnable transfer function $\transferfunsh_k$. Our contributed method composes the Gaussians in two modes---\textit{shadowed} and \textit{unshadowed}. The \textit{shadowed} model reconstructs additional shadows (see~\cref{fig:teaser}) on top of the unshadowed model thanks to our proposed use of a radiance transfer function. The Gaussians are splatted~\cite{kerbl20233dgaussiansplatting,zhang2024gaussian} to render the output image in a novel view and light. 
    }
    \label{fig:scheme}
\vspace{-1.1em}
\end{figure*}

\section{{Method}}

\label{sec:method}

\subsection{Preliminaries on Radiance Transfer}
\label{subsec:preliminaries}
The rendering equation, in its simplified form~\cite{green2003grittydetail}, is an integral function that represents light $L(\mathbf{x}, \direction_o)$ exiting point $\mathbf{x}$ along the vector $\direction_o$:
\begin{equation}
\label{eq:brdf}
L(\bx, \direction_o)\!=\!\int_{s} f_r(x, \direction_o, \direction_i) L_i(\bx, \direction_i) \transferfun(\bx, \direction_i) d\direction_i
\end{equation}
where $f_r(\cdot)$ is a BRDF function, $L_i(\cdot)$ an incoming light along the vector $\direction_i$, and $\transferfun(\cdot)$ is a radiance transfer function. Intuitively, $f_r(\cdot)$ represents the surface material, $L_i(\cdot)$ represents the intensity and color of the illumination, and $\transferfun(\cdot)$ is a term that takes into account shadows or light reflections from other surfaces. Depending on the formulation of those functions, the rendering equation can range from a straightforward and inaccurate light model to a highly complex and accurate one.

\paragraph{Unshadowed model} One example of a reflection model that can be represented with \cref{eq:brdf} is the diffuse surface reflection model, also known as \textit{dot product lighting}. A diffuse BRDF reflects light uniformly, making the lighting view-independent and simplifying the BRDF as follows:
\begin{equation}
\label{eq:dotillum}
L_{D}(\bx) = \frac{\rho(\bx)}{\pi} \int_{s} L_i(\bx, \direction_i) \max(\mathbf{n}(\bx) \cdot \direction_i, 0) d\direction_i
\end{equation}
where $\rho(\cdot)$ is the surface albedo, $\mathbf{n}(\bx)$ a surface normal at the point $x$. Shadows are neglected. 

The incoming light $\light_i(\bx, \direction_i)$ can be represented in several ways. In this work, we assume that the scene is illuminated with an \textbf{omnidirectional environment map} that is parametrized using spherical harmonics (SH) of degree $n$ with $(n{+}1)^2$ coefficients. Because the environment map is positioned infinitely far from the scene, the light is position-independent, and thus, the rendering equation is further simplified:
\begin{equation}
\label{eq:unshadowed}
\light_{U}(\bx) = \frac{\rho(\bx)}{\pi} \int_{s} \light_i(\direction_i) \max(\mathbf{n}(\bx) \cdot \direction_i, 0) d\direction_i
\end{equation}
With illumination parametrized with SH, we can evaluate the integral in~\cref{eq:unshadowed} using a closed-form solution from Eq.~(12) in \cite{ramamoorthi2001envmap}. From this point onward, we refer to rendering with \cref{eq:unshadowed} as \textit{unshadowed}.

\paragraph{Shadowed model} In addition to the \textit{unshadowed} lighting model, we propose a \textit{shadowed} model, where $\transferfun(\bx, \direction_i)$ is parameterized using spherical harmonics (SH) and learned from training data. In $\transferfun(\bx, \direction_i)$, SH represents a spherical signal that quantifies the light arriving from each direction of the environment map to an associated point in space. The \textit{shadowed} model is derived by replacing the dot product term in~\cref{eq:unshadowed}:
\begin{equation}
\label{eq:shadowed}
L_{S}(\bx) = \frac{\rho(\bx)}{\pi} \int_{s} L_i(\direction_i) \transferfun(\bx, \direction_i) d\direction_i.
\end{equation}
In addition to modeling shadows, this approach also has the potential to model the interreflection of light between objects in the scene. 

Using SH of the same degree for both the environment map and transfer function allows efficient evaluation of the rendering equation \cref{eq:shadowed}. A key SH property simplifies the integral of two SH-based functions to a dot product of their coefficients, thanks to SH orthogonality. With this property \cref{eq:shadowed} can be re-written as:
\begin{equation}
\label{eq:shadowed_sh}
L_{S}(\bx) = \frac{\rho(\bx)}{\pi} \lightsh \cdot \transferfunsh,
\end{equation}
where $\lightsh \in R^{(n+1)^2}$ are the SH coefficients of $L_i(\direction_i)$ and $\transferfunsh \in R^{(n+1)^2}$ are the SH coefficients of $D(\bx, \direction_i)$. Please see~\cite{slomp2006gentle, green2003grittydetail} for derivation.
This property is commonly used in real-time rendering where the radiance transfer function is pre-computed and only~\cref{eq:shadowed_sh} is evaluated at runtime.

\subsection{\ours}

\ours{} creates a 3D representation of a relightable model using 2D Gaussians~\cite{huang20242dgs} from $c\leq C$ images taken \textit{in-the-wild}  $\{\mathcal{I}_c\}_{c=1}^C$ with associated calibrated cameras~$\{\camera_c\}_{c=1}^C$. Our goal is to find Gaussian parameters $\mathcal{G}{=}\{\translation_k,\rotation_k,\scaling_k,\opacity_k, \albedo_k,\transferfunsh_k\}^K_{k=1}$  that  after the rasterization~\cite{kerbl20233dgaussiansplatting} recreate those images. We optimize Gaussians by minimizing the objective:
\begin{equation}
    \argmin_{\gaussians, \latents,\params} \mathbb{E}_{\camera_c\sim \{\camera_c\}} \underbrace{\loss{rgb}\!\left(\splatting(\camera_c\;|\;\gaussians, \latents, \params), \image_c\right)}_{\text{\cref{subsec:reconstruction}}}  + \underbrace{\mathcal{R}(\gaussians)}_{\text{\cref{subsec:constraints}}},
\end{equation}
where $\latents{=}\{\embedding_c\}_{c=1}^C$ is a set of scene-dependent, learnable environment embeddings, $\loss{rgb}$ is a photometric objective that compares the rendered image from an operator $\mathcal{S}(\cdot)$~(\cref{eq:splatting-op}), and $\mathcal{R}$ are additional regularization terms. In contrast to 2DGS~\cite{huang20242dgs}, for each Gaussian we model the base color $\albedo$ as diffuse\footnote{As per~\cref{subsec:preliminaries}, view-dependent effects are not modeled in diffuse reflections.}, and introduce SH coefficients for the transfer function $\transferfunsh$\footnote{These coefficients correspond to a single channel in practice.}. 2DGS provides smooth normals that make relighting possible. 

In what follows, we drop the dependence of functional forms on the positions $\bx$ we introduced in~\cref{subsec:preliminaries} for brevity.

\paragraph{Relighting} To handle the diverse lighting conditions in \textit{in-the-wild} images, we associate each training image with a learnable latent code $\embedding_\sceneindex$ that encodes its lighting conditions. Using this embedding, we predict the environment map coefficients via an MLP:

\begin{equation} \label{eq:lightsh} \lightsh_\sceneindex = \text{MLP}(\embedding_\sceneindex | \params), \end{equation}

where $\lightsh_\sceneindex \in \mathbb{R}^{3 \times (n+1)^2}$ represents the SH coefficients of the environment map, and $n{=}2$ is the SH degree. As shown in~\cite{ramamoorthi2001envmap}, second-order SH is sufficient to approximate environment lighting in many scenarios.

The predicted illumination is used in the rendering process in one of two ways: \textit{unshadowed} and \textit{shadowed}. Those two approaches correspond to~\cref{eq:unshadowed} and~\cref{eq:shadowed_sh} respectively, and are described below.

\paragraph{Unshadowed model}
For the unshadowed scenario, we follow~\cref{eq:unshadowed}, which integrates light over the hemisphere in the direction of the surface normal. The color $\radiance_k$, \textit{radiance}, for each Gaussian $\mathcal{G}_k$ given its normal $\normal_k$ and the illumination parameters $\mathbf{l}_\sceneindex$ equates to: 
\begin{equation}   
    \label{eq:unshadowed-radiance}
   \radiance_k = \albedo_k \odot \underbrace{\normal_k^t M(\lightsh_k) \normal_k}_{\text{unshadowed irradiance}},
\end{equation}
where $M$ is a $4{\times}4$ matrix derived from the SH parameters of the environment map. It is the closed form solution of the integral in \cref{eq:unshadowed}, please see Eq. (12) in \cite{ramamoorthi2001envmap} for details.

This simple yet effective model already imbues the model with relighting capabilities. However, as described in~\cref{fig:shadowed_unshadowed} it does not capture shadows correctly, limiting the output's fidelity. 

\def\scale{1.0}
\def\distbelow{2.5em}
\def\distabove{2.5em}
\def\height{256px}

\newcommand{\shadowedoverpicone}{
    \begin{overpic}[abs,height=\height]{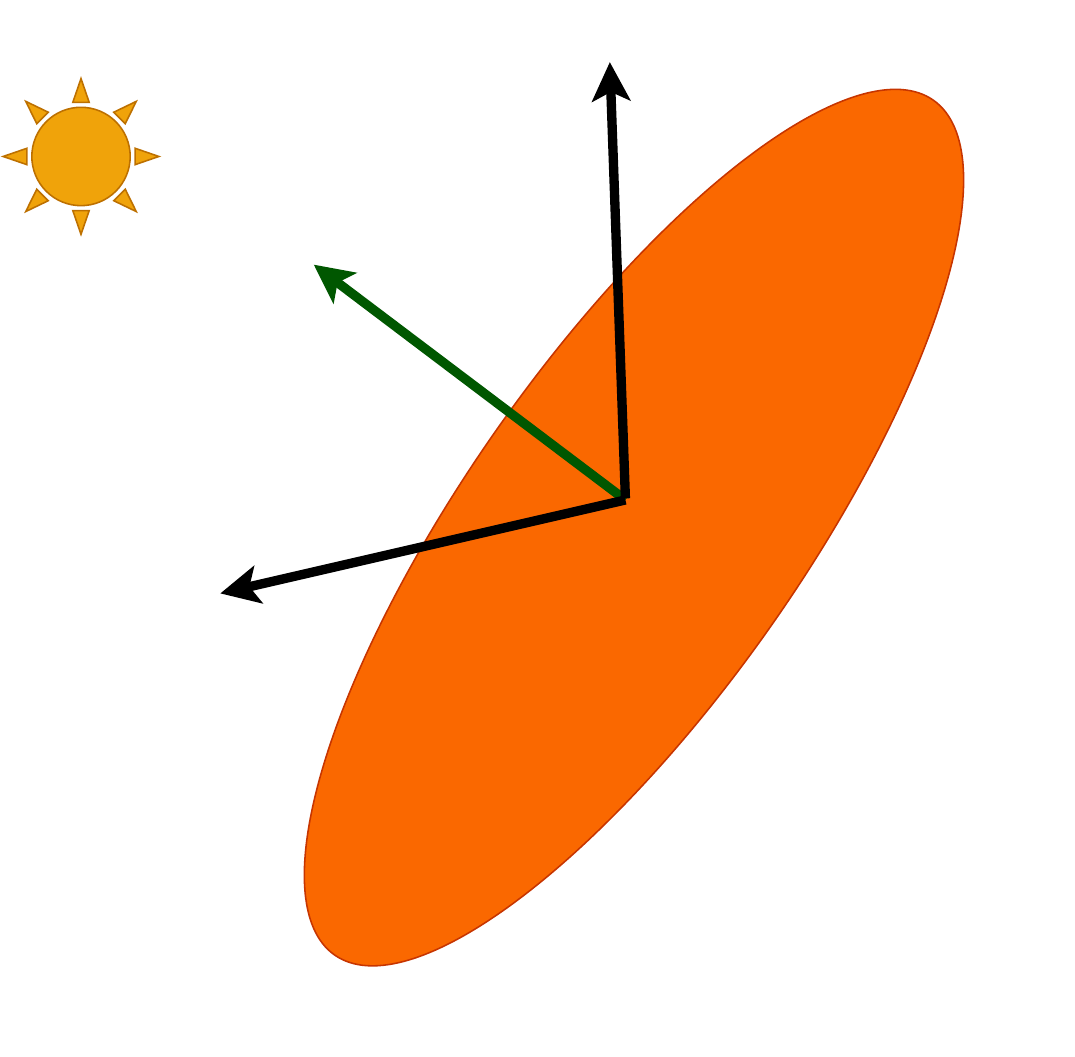}
        \put(90,200){$\normal$}
        \put(155,240){$\direction_1$}
        \put(65,130){$\direction_2$}
    \end{overpic}
}

\newcommand{\shadowedoverpictwo}{
    \begin{overpic}[abs,height=\height]{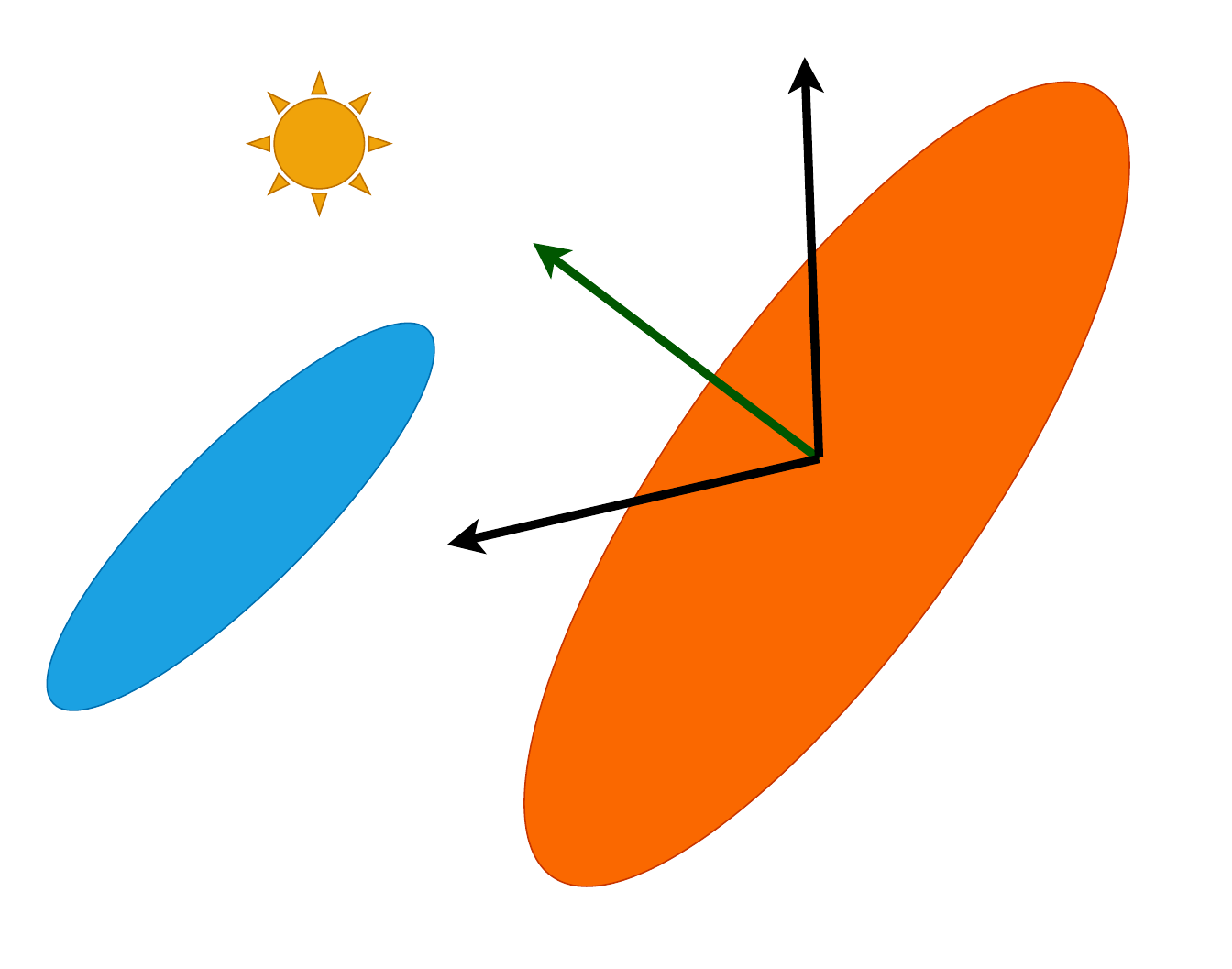}
        \put(155,200){$\normal$}
        \put(220,240){$\direction_1$}
        \put(130,130){$\direction_2$}
    \end{overpic}
}

\newcommand{\shadowedoverpicthree}{
    \begin{overpic}[abs,height=\height]{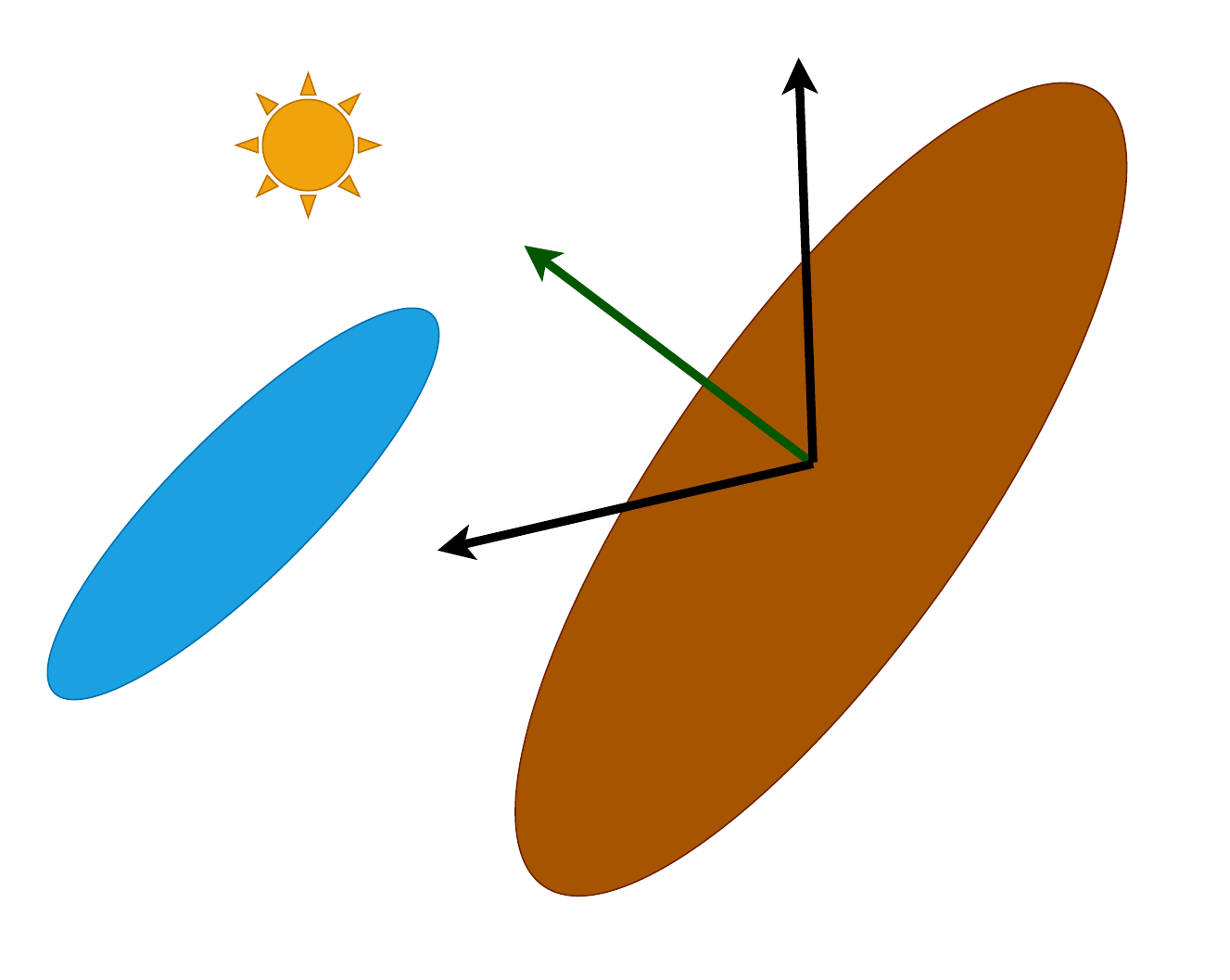}
        \put(155,200){$\normal$}
        \put(220,240){$\direction_1$}
        \put(130,130){$\direction_2$}
    \end{overpic}
}
\newcommand{\shadowedunshadowed}{
\centering
\resizebox{\linewidth}{!}{
    \begin{tikzpicture}[
        >=stealth',
        overlay/.style={
          anchor=south west, 
          draw=black,
          rectangle, 
          line width=0.8pt,
          outer sep=0,
          inner sep=0,
        },
        font=\Huge
    ]
    \matrix[matrix of nodes, column sep=1pt, row sep=0pt, ampersand replacement=\&, inner sep=0, outer sep=0, font=\Huge] (shadowedunshadowed) {
        \shadowedoverpicone{} \&
        \shadowedoverpictwo{} \&
        \shadowedoverpicthree{} \\
    }; 
    \node[below=\distbelow of shadowedunshadowed-1-1.south, anchor=base, scale=\scale] {
    $\begin{aligned}
        \transferfunsh(\direction_1)&{=}\max(0, \cos(\normal, \direction_1)) \\
        \transferfunsh(\direction_2)&{=}\max(0, \cos(\normal, \direction_2))
    \end{aligned}$
    };
    
    \node[below=\distbelow of shadowedunshadowed-1-2.south, anchor=base, scale=\scale] {
    $\begin{aligned}
        \transferfunsh(\direction_1)&{=}\max(0, \cos(\normal, \direction_1)) \\
        \transferfunsh(\direction_2)&{=}\max(0, \cos(\normal, \direction_2))
    \end{aligned}$
    };
    
    \node[below=\distbelow of shadowedunshadowed-1-3.south, anchor=base, scale=\scale] {
    $\begin{aligned}
        \transferfunsh(\direction_1)&{=}\max(0, \cos(\normal, \direction_1)) \\
        \transferfunsh(\direction_2)&{<}\max(0, \cos(\normal, \direction_2))
    \end{aligned}$
    };

    \node[above=\distabove of shadowedunshadowed-1-1.north, anchor=base, scale=\scale, align=center]{
        $\radiance = \shadowedradiance$ \\
        (\cref{eq:unshadowed-radiance} \& \cref{eq:shadowed-radiance})
    };

    \node[above=\distabove of shadowedunshadowed-1-2.north, anchor=base, scale=\scale, align=center]{
        $\radiance$ does not incorporate \\
        the light blocker
    };
    
    \node[above=\distabove of shadowedunshadowed-1-3.north, anchor=base, scale=\scale, align=center]{
        $\shadowedradiance$ incorporates \\
        the light blocker
    };
    
\end{tikzpicture}
}
}

\begin{figure}[!t]
    \centering
    \shadowedunshadowed
    \caption{%
            Unshadowed $\radiance$ (\cref{eq:unshadowed-radiance}) and shadowed $\shadowedradiance$ (\cref{eq:shadowed-radiance}) may give the same output color if a Gaussian is fully exposed to the environment light. In the case of any occluder, $\radiance$ does not handle, and the color does not change. However, our proposed $\shadowedradiance$ properly reacts to the occluder and makes the output color darker.
    }
    
    \label{fig:shadowed_unshadowed}
    \vspace{-1em}
\end{figure}

\paragraph{Shadowed model}
To effectively capture shadows in the model, we redefine the output color of a Gaussian as $\shadowedradiance_k$, a function of learnable radiance transfer function $\transferfun_k$ parametrized by spherical harmonics $\transferfunsh_k \in \mathbb{R}^{(n+1)^2}$, light $\lightsh_c$ and albedo $\albedo_k$. Using a learned radiance transfer function (instead of fixing it to capture light from the hemisphere above the normal as we do in \textit{unshadowed}) allows for creating shadows, as described in \cref{subsec:preliminaries}. Overall, following \cref{eq:shadowed_sh}, the output shadowed color or \textit{radiance} reduces to:
\begin{equation}
    \label{eq:shadowed-radiance}
    \shadowedradiance_{k}   = \albedo_k \odot \underbrace{\sum_{i=1}^{(n+1)^2}{\mathbf{l}_c^i \cdot \mathbf{d}_{k}^i}}_{\text{shadowed irradiance}},
\end{equation}
As we show later in the experiments, the addition of shadows leads to more accurate relighting. Additionally, it does not require learnable MLP to reconstruct shadows at the inference stage, differentiating it from NeRF-OSR~\cite{rudnev2022nerfosr} and making our approach applicable to rendering engines directly.

\def\spysize{24px}
\def\offset{36px}
\def\height{64px}
\newcommand{\inpscene}[1]{\includegraphics[height=\height]{images/sep_images/relighting/#1.png}}

\newcommand{\relighting}{
\centering
\resizebox{\linewidth}{!}{
    \begin{tikzpicture}[
        >=stealth',
        overlay/.style={
          anchor=south west, 
          draw=black,
          rectangle, 
          line width=0.8pt,
          outer sep=0,
          inner sep=0,
        },
        font=\Huge
    ]
    \matrix[
        matrix of nodes, 
        column sep=0pt,
        row sep=0pt,
        ampersand replacement=\&,
        inner sep=0,
        outer sep=0
    ] (pictures) {
        \inpscene{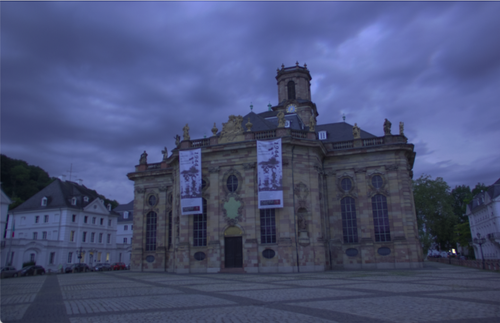} \&
        \inpscene{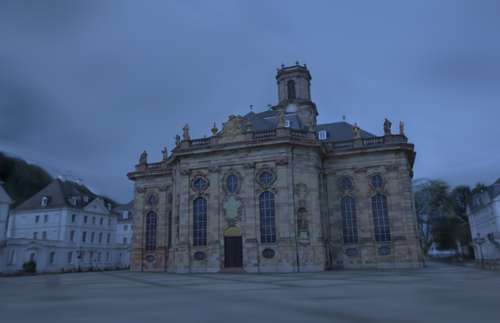} \&
        \inpscene{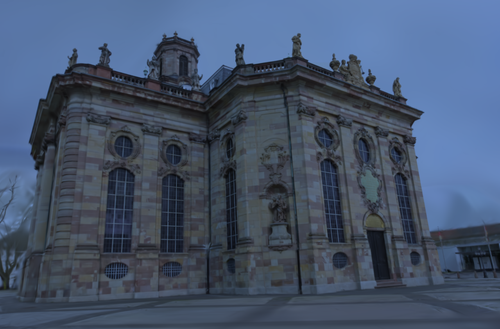} \\
        \inpscene{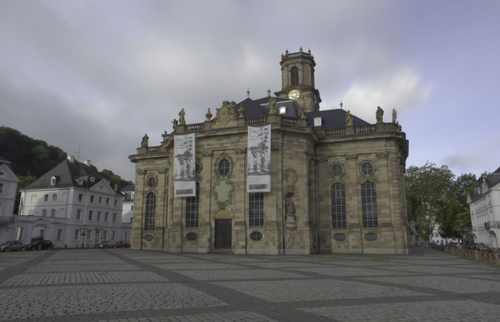} \&
        \inpscene{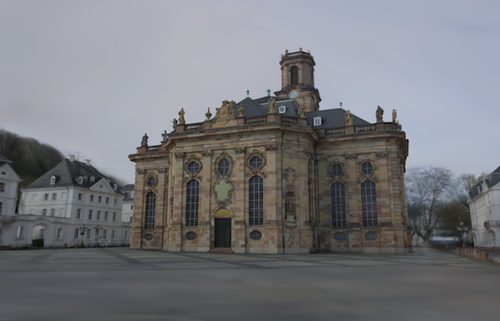} \&
        \inpscene{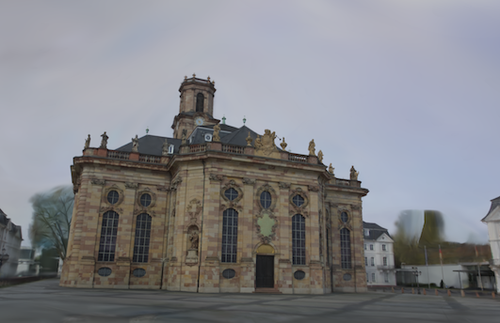} \\
        \inpscene{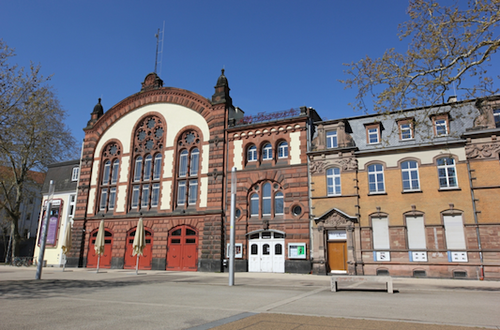} \&
        \inpscene{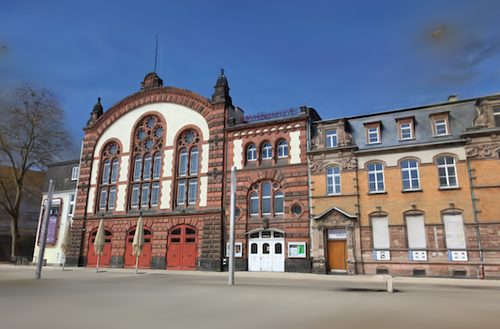} \&
        \inpscene{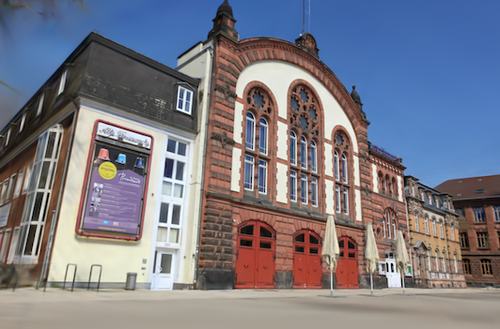} \\
        \inpscene{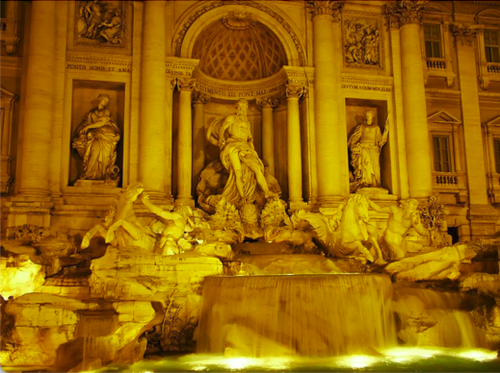} \&
        \inpscene{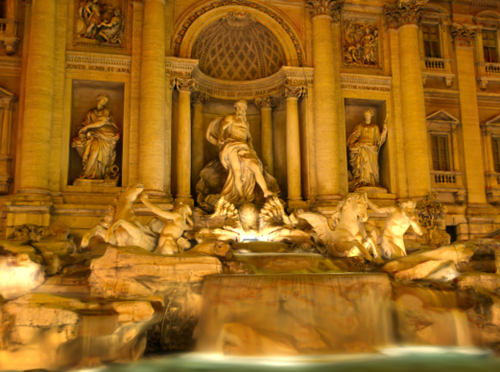} \&
        \inpscene{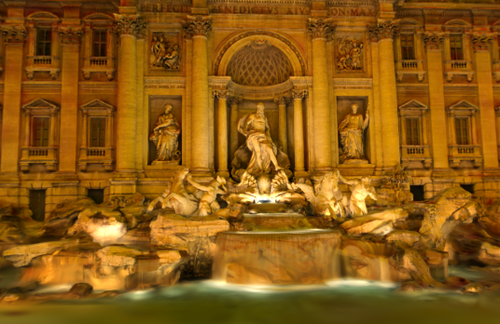} \\
    }; 
    \matrix[
        matrix of nodes, 
        column sep=0pt,
        row sep=0pt,
        ampersand replacement=\&,
        inner sep=0,
        outer sep=0,
        right=of pictures
    ] (chair) {
        \inpscene{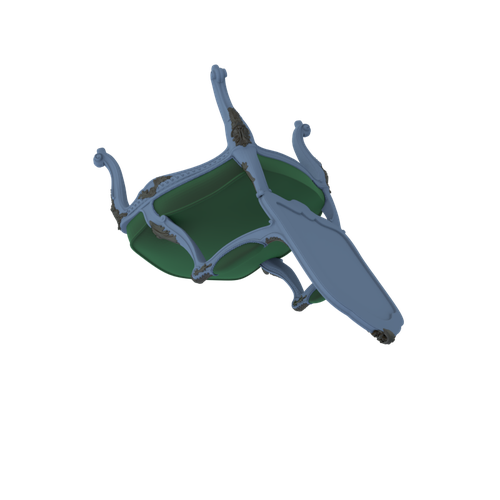} \\
        \inpscene{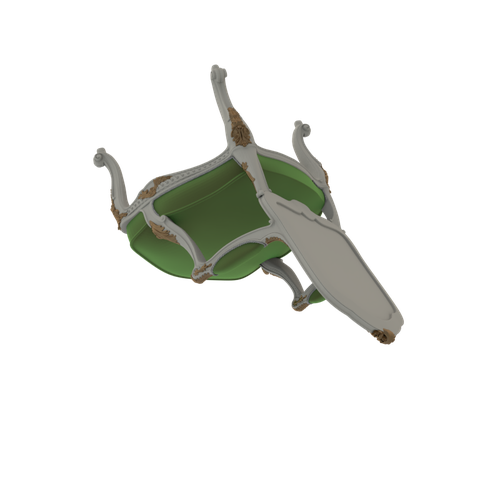} \\
        \inpscene{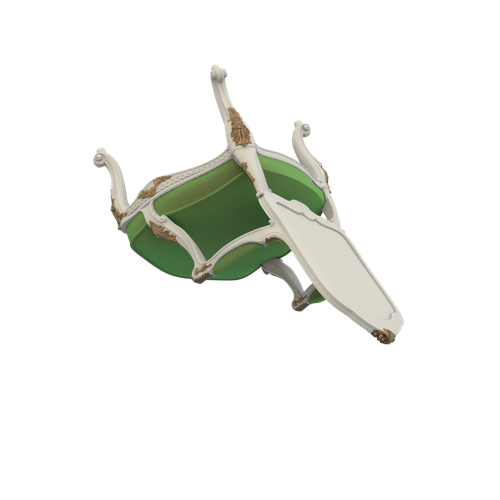} \\
        \inpscene{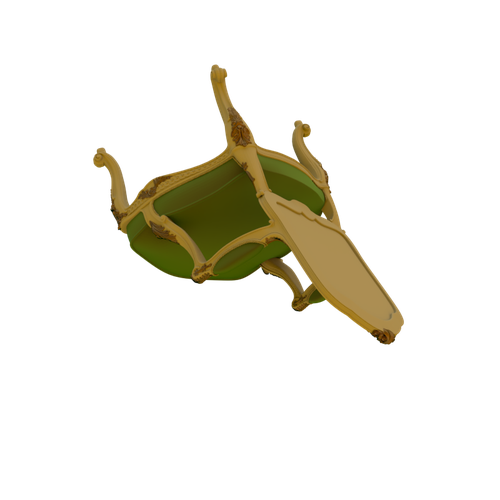} \\
    }; 
    \matrix[
        matrix of nodes, 
        column sep=0pt,
        row sep=0pt,
        ampersand replacement=\&,
        inner sep=0,
        outer sep=0,
        right=of chair
    ] (chair) {
        \inpscene{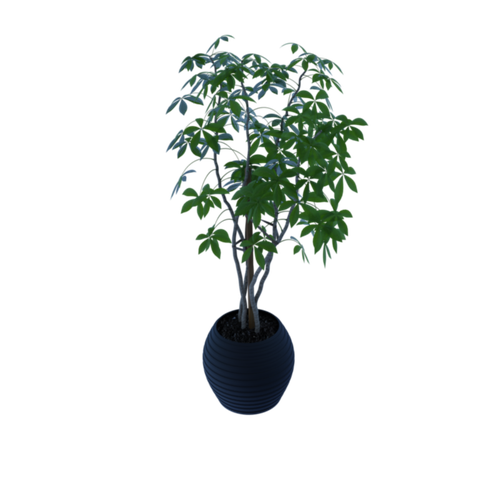} \\
        \inpscene{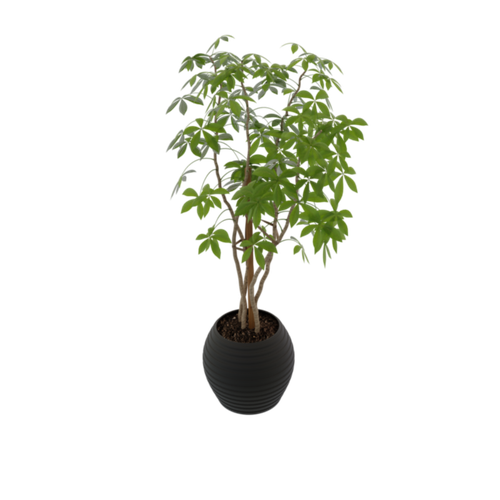} \\
        \inpscene{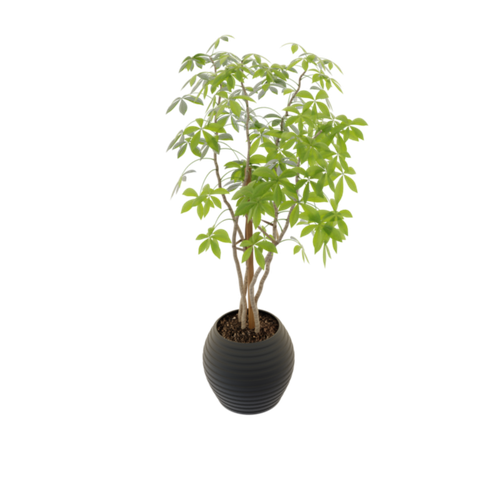} \\
        \inpscene{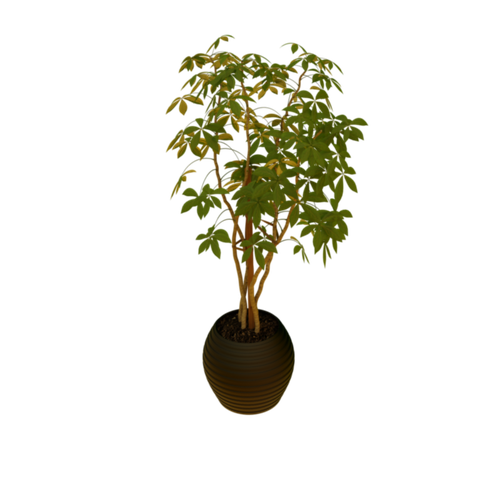} \\
    }; 
    
\end{tikzpicture}
}
}

\begin{figure}[!t]
    \centering
    \begin{overpic}[percent,width=\linewidth]{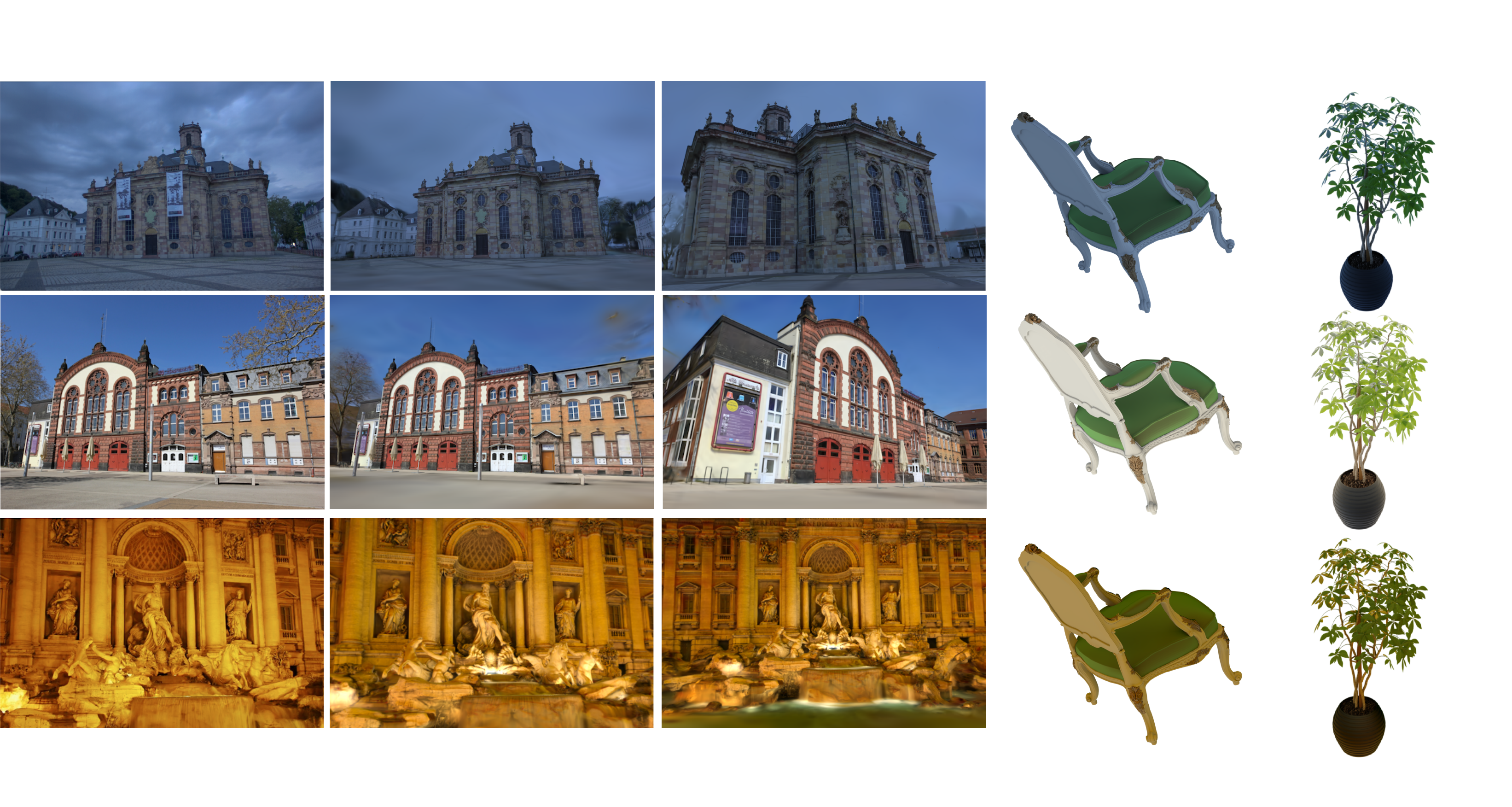}
    \put(2,49.5) {\footnotesize Target Image}
    \put(22.5,49.5) {\footnotesize Reconstruction}
    \put(46.5,49.5) {\footnotesize Novel View}
    \put(71.2,49.5) {\footnotesize Relit Objects}
    \end{overpic}
    \vspace{-3em}
    \caption{\textbf{Scene reconstruction and relightning --} Reconstruction and relighting capabilities of \ours{}. \change{\ours{} reproduces sharp and clean landmarks, and the learned environment lighting enables accurate scene relighting. We use learned environment maps to relight the scene from novel viewpoints and then relight arbitrary objects within a graphics engine.}}
    \label{fig:qualitative_ours}
    \vspace{-1em}
\end{figure}

\subsection{Physical constraints}
\label{subsec:constraints}
The regularizations proposed in 2DGS~\cite{huang20242dgs} keep the Gaussians close to the surface and smooth locally, which is crucial in our relighting scenario. Aside from them, we propose new loss terms based on the physical light properties that restrict the optimization from achieving degenerate, \textit{non-relightable} cases. We restrict radiance transfer $\transferfun_k$ function to remain within the range of 0 to 1, where 0 indicates complete shadowing and 1 signifies full exposure~to~lighting:
\begin{equation}
\begin{split}
    \loss{0-1}&=\mathbb{E}_k\mathbb{E}_{\direction_i} [\|\max(\transferfun_k(\direction_i), 1)-1\|^2_2 \\
    &\qquad\quad+ \|\min(\transferfun_k(\direction_i), 0)\|^2_2],
\end{split}
\end{equation}
and allow the environment light to remain in the $\mathbb{R}_{+}$ domain:
\begin{equation}
    \loss{+} = \mathbb{E}_k\mathbb{E}_{\direction_i}\|\min(\light_\sceneindex(\direction_i), 0)\|^2_2,
\end{equation}
which allows the environment light to brighten the scene arbitrarily.  

The shadowed radiance transfer should remain close to the unshadowed version. If not, the shadowed version might include light from any direction, resulting in degenerate solutions and incorrect relighting. We visualize the shadowed and unshadowed transfer functions in \cref{fig:shadowed_unshadowed}. To address this issue, we propose the following loss function:
\begin{equation}
    \loss{\moon{8}$\leftrightarrow$\moon{16}} = \mathbb{E}_k\mathbb{E}_{\direction_i}\|\max(\normal_k \cdot \direction_i, 0) - \transferfun_k(\direction_i)\|^2_2,
\end{equation}

\begin{table*}[t]
\centering
\caption{\textbf{Quantitative results --} Comparison between our \ours{} and selected baselines for two different . We report the reconstruction quality regarding PSNR, MSE, MAE, SSIM on full and 4x downsampled image resolutions.
\textit{u/s} denotes using upsampled, \textit{d/s} downsampled images for the evaluation, and the last delimited area presents the ablation study on downsampled data. 
\change{We denote NeRF-OSR \cite{rudnev2022nerfosr} results reproduced by FEGR~\cite{wang2023fegr} with  *.} We use $\dagger$ to further annotate our approach where we remove loss terms $\loss{\shadowed$\leftrightarrow$\unshadowed}$, $\loss{rec}(\unshadowed)$ from the second training stage.
In $\ddagger$, we omit the first training stage. 
\change{Compared to the baselines, \ours{} achieves reconstructions of high fidelity. It reliably produces smooth surfaces and sharp edges, reflected in its high SSIM values. Additionally, our proposed components either enhance reconstruction or preserve physical accuracy without negatively impacting the results.
\kk{Please note, that NeuSky~\cite{gardner2023neusky} is a concurrent work, published prior to the WACV's deadline at ECCV 2024.}
}
}

\setlength{\tabcolsep}{4pt}  
\resizebox{\linewidth}{!}{
\begin{tabular}{lcccccccccccc}
\toprule
\multirow{2}[2]{*}{Method}& \multicolumn{4}{c}{Landwehrplatz} & \multicolumn{4}{c}{Ludwigskirche} & \multicolumn{4}{c}{Staatstheater} \\
\cmidrule(lr){2-5} \cmidrule(lr){6-9} \cmidrule(lr){10-13}
 & PSNR $\uparrow$ & MSE $\downarrow$ & MAE $\downarrow$ & SSIM $\uparrow$ & PSNR $\uparrow$ & MSE $\downarrow$ & MAE $\downarrow$ & SSIM $\uparrow$ & PSNR $\uparrow$ & MSE $\downarrow$ & MAE $\downarrow$ & SSIM $\uparrow$ \\
\midrule
Yu et al. $_{u/s}$\cite{yu2020self} & 15.17 & 0.033 & 0.133 & 0.376 & 17.87 & 0.017 & 0.097 & 0.378 & 15.28 & 0.032 & 0.138 & 0.385 \\
Philip et al. \cite{philip2019multi} & 12.28 & 0.062 & 0.179 & 0.319 & 16.63 & 0.023 & 0.113 & 0.367 & 12.34 & 0.065 & 0.200 & 0.272 \\
NeRF-OSR \cite{rudnev2022nerfosr} & 16.65 & 0.024 & 0.114 & 0.501 & 18.72 & 0.014 & 0.090 & 0.468 & 15.43 & 0.029 & 0.133 & 0.517 \\
NeRF-OSR* \cite{rudnev2022nerfosr} & 15.66 & 0.029 & - & - & 19.34 & 0.012 & - & - & 16.35 & 0.027 & - & - \\
SR-TensoRF \cite{chang2024srtensorf} & 16.74 & 0.024 & 0.093 & 0.653 & 17.30 & 0.021 & 0.096 & 0.542 & 15.43 & 0.030 & 0.111 & 0.632 \\
FEGR \cite{wang2023fegr} & 17.57 & 0.018 & - & - & 21.53 & 0.007 & - & - & 17.00 & 0.023 & - & - \\
SOL-NeRF \cite{solnerf} & 17.58 & 0.028 & - & 0.618 & 21.23 & 0.008 & - & 0.749 & 18.18 & 0.019 & - & 0.680 \\
NeuSky \cite{gardner2023neusky} & 18.31 & 0.016 & - & - & 22.50 & 0.005 & - & - & 16.66 & 0.023 & - & - \\
\textbf{Ours} & 18.01 & 0.017 & 0.096 & 0.778 & 19.59 & 0.012 & 0.085 & 0.700 & 17.02 & 0.021 & 0.107 & 0.729 \\
\midrule
Yu et al. \cite{yu2020self} & 15.84 & 0.028 & 0.123 & 0.392 & 18.71 & 0.014 & 0.088 & 0.400 & 15.43 & 0.031 & 0.136 & 0.363 \\
Philip et al. $_{d/s}$ \cite{philip2019multi} & 12.85 & 0.054 & 0.169 & 0.164 & 17.37 & 0.019 & 0.105 & 0.429 & 11.85 & 0.070 & 0.210 & 0.184 \\
NeRF-OSR $_{d/s}$ \cite{rudnev2022nerfosr}  & 17.38 & 0.021 & 0.106 & 0.576 & 19.86 & 0.011 & 0.080 & 0.626 & 15.83 & 0.026 & 0.128 & 0.556 \\
\textbf{Ours}~$_{d/s}$ & 18.40 & 0.016 & 0.094 & 0.746 & 20.13 & 0.011 & 0.080 & 0.727 & 17.24 & 0.020 & 0.105 & 0.715 \\
\midrule
$\text{\textbf{Ours}}~\dagger$& 15.03 & 0.034 & 0.139 & 0.58 & 19.34 & 0.015 & 0.094 & 0.693 & 16.09 & 0.028 & 0.124 & 0.665 \\
$\text{\textbf{Ours}}~\ddagger$  & 17.59 & 0.019 & 0.100 & 0.733 & 19.05 & 0.016 & 0.097 & 0.680 & 16.83 & 0.022 & 0.110 & 0.694 \\
$\text{\textbf{Ours}}\setminus\loss{0-1}$ & 18.30 & 0.016 & 0.095 & 0.744 & 20.15 & 0.010 & 0.080 & 0.734 & 17.25 & 0.020 & 0.105 & 0.712 \\
$\text{\textbf{Ours}}\setminus\loss{+}$ & 17.35 & 0.020 & 0.104 & 0.728 & 20.17 & 0.012 & 0.081 & 0.729 & 17.10 & 0.020 & 0.106 & 0.703 \\

\bottomrule
\end{tabular}
}
\vspace{-1em}
\label{tab:results_osr_eval}
\end{table*}

\begin{table}[t]
\centering
\caption{\textbf{Performance comparison --} \kk{Training time and inference speed comparison between the baselines and our \ours{}.}}
\vspace{-0.2em}
\begin{tabular}{lccc}
\toprule
\textbf{Method} & \textbf{Training time} & \textbf{FPS}  \\
\midrule
NeRF-OSR \cite{rudnev2022nerfosr} & 31h & 0.003 \\
NeuSky \cite{gardner2023neusky} & 14h & 0.004 \\
\textbf{Ours} & 1h 20min & 20.7  \\
\bottomrule
\end{tabular}
\vspace{-1.2em}
\label{tab:performance}
\end{table}

The applied transfer function inherently accounts for shadows and interreflections. To focus specifically on modeling shadows and restrict the use of~\cref{eq:shadowed-radiance} for other cases, we impose a loss function ensuring that shadowed radiance should not be brighter than unshadowed one:
\begin{equation}
    \loss{\moon{8}} = \mathbb{E}_k\mathbb{E}_{\direction_i}\| \max(\transferfun_k(\direction_i) -\max(\normal_k \cdot \direction_i, 0), 0) \|^2_2,
\end{equation} 
Those losses are weighted with scalars $\{\lambda_{1,\dots,4}\}$ fixed across experiments and contribute to our regularization term:
\begin{equation}
    \mathcal{R}(\gaussians) = \lossweight{1}\loss{0-1} + \lossweight{2}\loss{+} + \lossweight{3}\loss{\moon{8}$\leftrightarrow$\moon{16}} +
    \lossweight{4}\loss{\moon{8}}
\end{equation}
Calculating it exactly requires us to compute the expectation over the hemisphere $\mathbb{S}^2$. Instead, we approximate the expectations over directions $\direction_i$ with a Monte Carlo estimator by randomly sampling the SH lobe with $N$ samples at each training step.

\subsection{Reconstruction}
\label{subsec:reconstruction}
We render images using the splatting algorithm $\splatting(\cdot)$ proposed in 2DGS~\cite{huang20242dgs}. We compare the rendered images with ground-truth  $\{\image_c\}$ taken with $\{\camera_c\}$ cameras. Our method builds on 2DGS~\cite{huang20242dgs} and therefore our reconstruction loss $\loss{rgb}$ follows the following term:
\begin{align}
    \loss{rgb} &= \lossweight{rec}(\shadowed)\loss{rec}(\shadowed) + \lossweight{rec}(\unshadowed)\loss{rec}(\unshadowed),\\
    \loss{rec}(\{\shadowed, \unshadowed\}) &= \loss{1}(\{\shadowed, \unshadowed\}) + \lossweight{}\loss{D-SSIM}(\{\shadowed, \unshadowed\}),
\end{align}
where the $\loss{1}$ is the $L_1$ loss comparing either the image rendered from our shadowed or unshadowed models and $\loss{D-SSIM}$ is a differentiable D-SSIM~\cite{ssim} further improving the quality. We use $\lossweight{}{=}0.2$ throughout all the experiments. Our proposed $\loss{rec}(\unshadowed)$ resembles a pretraining stage. 
As the more complex shadowed model lands in local minima if trained from scratch, we initiate the training with $\lossweight{rec}(\shadowed){=}0.0$ and $\lossweight{rec}(\unshadowed){=}1.0$. Once the simpler model converges, we switch $\lossweight{rec}(\shadowed){=}1.0$ and $\lossweight{rec}(\unshadowed)$ to a small value so as not to deteriorate the quality of the model. In short, the shadowed model explains the parts of an image with shadows, which the unshadowed could not with its simpler lighting model.

\def\height{64px}
\def\spysize{24px}
\def\offset{36px}
\def\scale{0.7}
\def\scaleleft{0.7}
\def\distabove{0.3em}
\def\distleft{0.0em}
\def\clipleft{2cm}
\def\clipright{2cm}
\renewcommand{\inputimage}[1]{\includegraphics[height=\height,trim={8px 0 8px 0}, clip]{images/sep_images/qual_relight_albedo_norm/#1}}
\renewcommand{\inputimageshadows}[1]{\includegraphics[height=\height,trim={8px 0 8px 0}, clip]{images/sep_images/qual_gt_shadows/#1}}

\newcommand{\newqualitative}{
\centering
\resizebox{\linewidth}{!}{
    \begin{tikzpicture}[
        >=stealth',
        title/.style={
            anchor=base,
            align=center,
            scale=\scale,
        },
        spy using outlines={circle, red, magnification=3, connect spies, size=\spysize}
    ]
    \matrix[matrix of nodes, column sep=0pt, row sep=0pt, ampersand replacement=\&, inner sep=0, outer sep=0, font=\Huge] (qualitative) {
        \inputimage{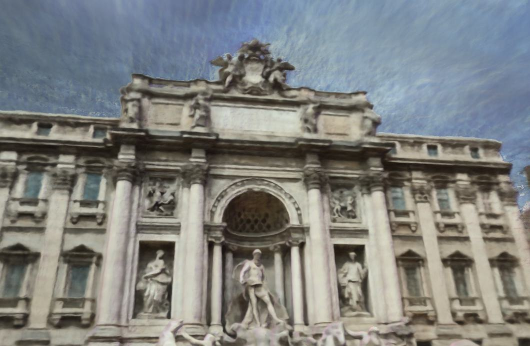} \&
        \inputimage{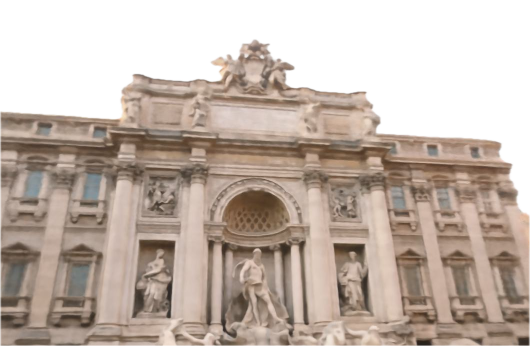} \&
        \inputimage{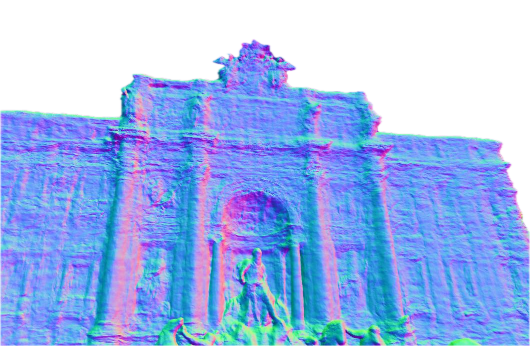} \&
        \inputimage{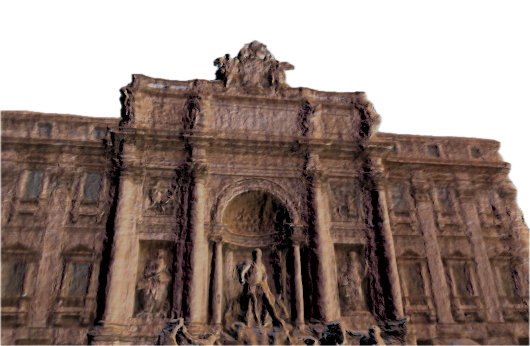} \\
        \inputimage{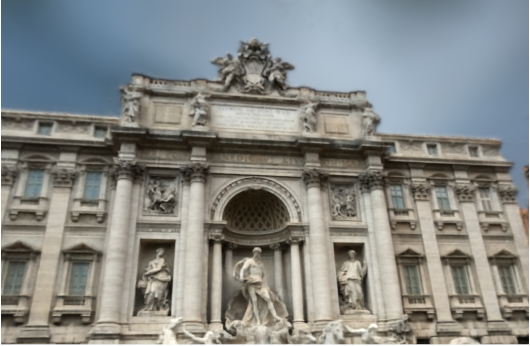} \&
        \inputimage{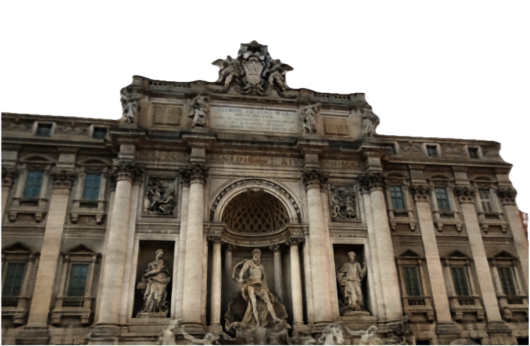} \&
        \inputimage{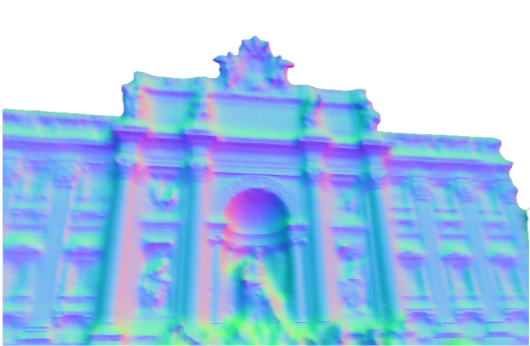} \&
        \inputimage{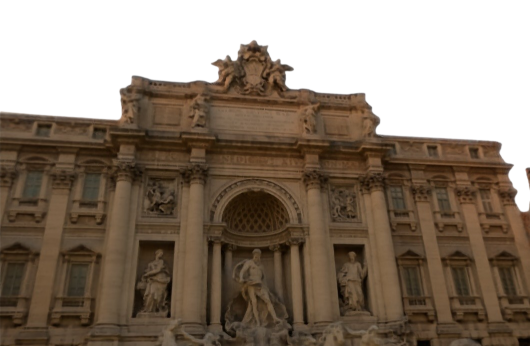} \\
        \inputimage{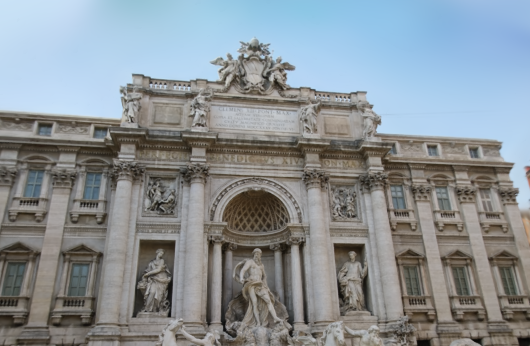} \&
        \inputimage{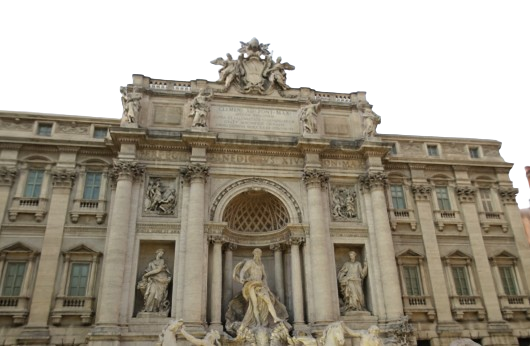} \&
        \inputimage{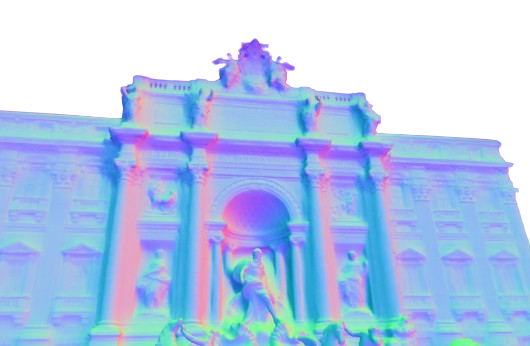} \&
        \inputimage{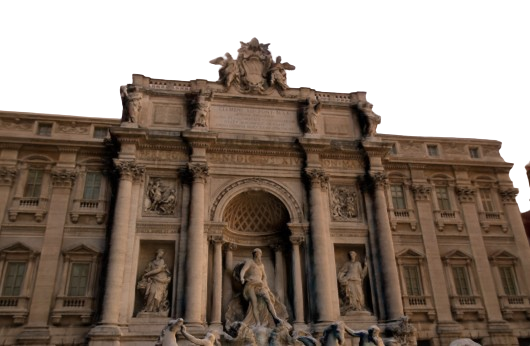} \\
    }; 
    
    \node[above=\distabove of qualitative-1-1.north, title] {Reconstruction};
    \node[above=\distabove of qualitative-1-2.north, title] {Albedo};
    \node[above=\distabove of qualitative-1-3.north, title] {Normals};
    \node[above=\distabove of qualitative-1-4.north, title] {Novel Lightning};

    \node[left=\distleft of qualitative-1-1.west, align=center,scale=\scaleleft]{\rotatebox{90}{NeRF-OSR~\cite{rudnev2022nerfosr}}};
    \node[left=\distleft of qualitative-2-1.west, align=center,scale=\scaleleft]{\rotatebox{90}{NeuSky~\cite{gardner2023neusky}}};
    \node[left=\distleft of qualitative-3-1.west, align=center,scale=\scaleleft]{\rotatebox{90}{\textbf{Ours}}};

    \foreach \n in {1,...,3}{
        \edef\temp{\noexpand\spy[anchor=center,magnification=3] on ([xshift=0.0em,yshift=-0.5em] qualitative-\n-1.center) in node[below left=3.75px and 3.75px] at (qualitative-\n-1.north east);}
        \temp
    }

    \foreach \n in {1,...,3}{
        \edef\temp{\noexpand\spy[anchor=center,magnification=3] on ([xshift=-1.0em,yshift=0.0em] qualitative-\n-2.center) in node[below left=3.75px and 3.75px] at (qualitative-\n-2.north east);}
        \temp
    }
    \foreach \n in {1,...,3}{
        \edef\temp{\noexpand\spy[anchor=center,magnification=3] on ([xshift=1.0em,yshift=0.5em] qualitative-\n-3.center) in node[below left=3.75px and 3.75px] at (qualitative-\n-3.north east);}
        \temp
    }
    \foreach \n in {1,...,3}{
        \edef\temp{\noexpand\spy[anchor=center,magnification=3] on ([xshift=-0.7em,yshift=-2em] qualitative-\n-4.center) in node[below left=3.75px and 3.75px] at (qualitative-\n-4.north east);}
        \temp
    }
\end{tikzpicture}
}
}

\begin{figure*}[!t]
    \centering
    \newqualitative
    \caption{\textbf{Qualitative comparison of albedo, normals, and relighting under similar lighting conditions on Trevi Fountain.} \ours{} produces albedo with fewer baked-in shadows, sharp normals, smooth surfaces, and more accurate novel lighting compared to the baselines. Results for NeuSky originally reported in \cite{gardner2023neusky}. Please, zoom in for details.} 
    \label{fig:qual_albedo_norm_relight}
    \vspace{-1em}
\end{figure*}

\def\spysize{24px}
\def\offset{36px}
\def\height{84px}
\newcommand{\inpscenenormals}[1]{\includegraphics[height=\height]{images/sep_images/normals/#1.png}}

\newcommand{\normalsshadows}{
\centering
\resizebox{\linewidth}{!}{
    \begin{tikzpicture}[
        >=stealth',
        anchor=base,
        align=center,
        spy using outlines={circle, red, magnification=2, connect spies, size=52px}
    ]
    \matrix[
        matrix of nodes, 
        column sep=0pt,
        row sep=0pt,
        ampersand replacement=\&,
        inner sep=0,
        outer sep=0
    ] (normalsshadows) {
        \inpscenenormals{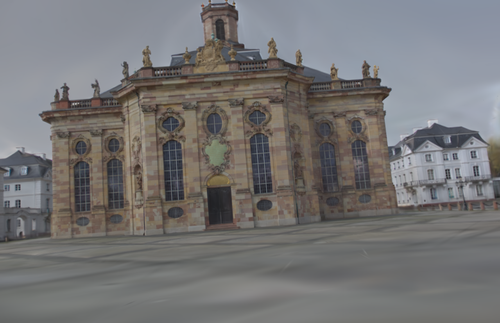} \&
        \inpscenenormals{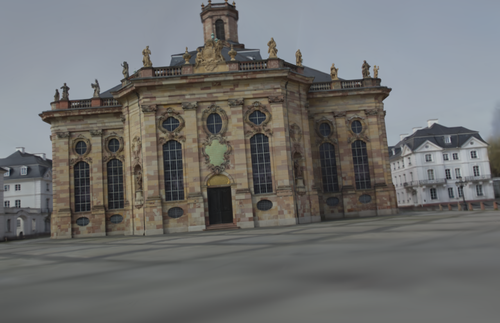} \\
        \inpscenenormals{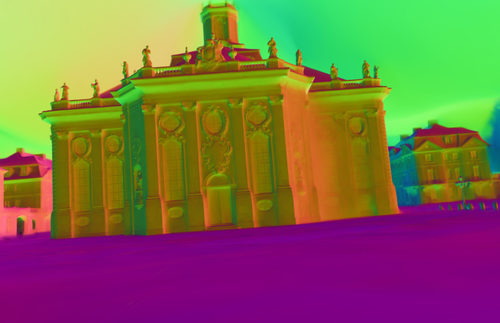} \&
        \inpscenenormals{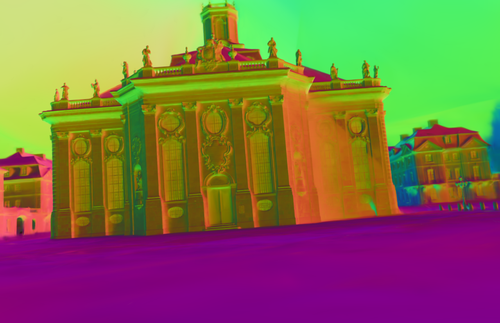} \\
    }; 

    \node[above=0.1em of normalsshadows-1-2.north] {Model without shadows $\unshadowed$};
    \node[above=0.1em of normalsshadows-1-1.north] {Shadowed model $\shadowed$};

    \spy[anchor=center] on ([xshift=2.5em,yshift=1.5em] normalsshadows-1-1.center) in node[above right=8px and 8px] at (normalsshadows-1-1.south west);
    \spy[anchor=center] on ([xshift=2.5em,yshift=1.5em] normalsshadows-1-2.center) in node[above right=8px and 8px] at (normalsshadows-1-2.south west);
    \spy[anchor=center] on ([xshift=2.5em,yshift=1.5em] normalsshadows-2-1.center) in node[above right=8px and 8px] at (normalsshadows-2-1.south west);
    \spy[anchor=center] on ([xshift=2.5em,yshift=1.5em] normalsshadows-2-2.center) in node[above right=8px and 8px] at (normalsshadows-2-2.south west);
\end{tikzpicture}
}
}

\begin{figure}[t]
    \centering
    \normalsshadows
    \caption{\textbf{Effects of shadowed training --} We show the comparison of \textbf{albedo} between the shadowed (\textit{left}) and unshadowed (\textit{right}) models. The albedo in the shadowed training is brighter with fewer shadows. The shadowed model recovers more accurate normals.}
    \label{fig:shadowed_fix_normals}
\vspace{-1em}
\end{figure}

\section{Experiments}
\label{sec:experiments}

\subsection{Datasets and baselines}
To evaluate our approach, we followed the protocol from NeRF-OSR~\cite{rudnev2022nerfosr} using ground truth environment maps. We use the official data split for Staatstheatert, Landwehrplatz, and Ludwigskirche. We use segmentation masks for test images provided in the OSR dataset and calculate MSE, MAE, SSIM and PSNR on masked regions only. We compare \ours{} against \change{several NeRF-based baselines\footnote{We include the concurrent NeuSky~\cite{gardner2023neusky} which has been published officially after the WACV deadline.} and TensoRF baseline}. 
We provide the implementation details in~\supplementary{}.

\subsection{Scene reconstruction and relightning}

 \kk{%
 We present the qualitative results in~\cref{tab:results_osr_eval} and quantitative in~\cref{fig:qual_albedo_norm_relight,fig:qualitative_ours}. As Yu~\etal~\cite{yu2020self} evaluates their model on downsampled images, we show the metric values on downsampled (\textit{d/s}), and upsampled (\textit{u/s}) to identify the quality differences. As we can see, \ours{} performs better or on par with the baselines. NeuSky~\cite{gardner2023neusky} is a concurrent work which models the environment maps and the sky using a prior, pretrained model.}

\def\scale{1.0}
\def\distabove{0.5em}
\def\height{64px}
\def\spysize{24px}
\def\offset{36px}
\newcommand{\inputimagerotate}[1]{\includegraphics[height=\height]{images/sep_images/rotating/#1.png}}

\newcommand{\rotatingfigure}{
\centering
\resizebox{\linewidth}{!}{
    \begin{tikzpicture}[
        >=stealth',
        spy using outlines={circle, red, magnification=3, connect spies, size=\spysize}
    ]
    \matrix[matrix of nodes, column sep=0pt, row sep=0pt, ampersand replacement=\&, inner sep=0, outer sep=0] (scene1) {
        \inputimagerotate{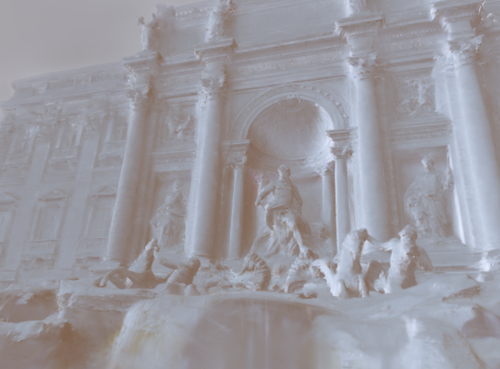} \&
        \inputimagerotate{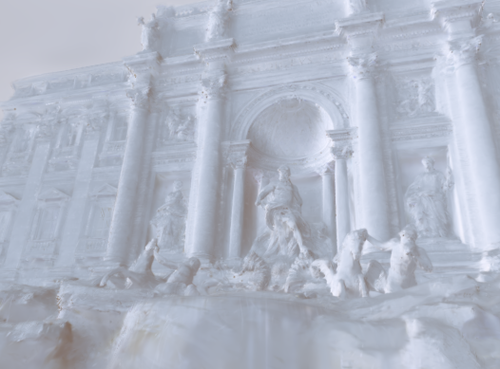} \&
        \inputimagerotate{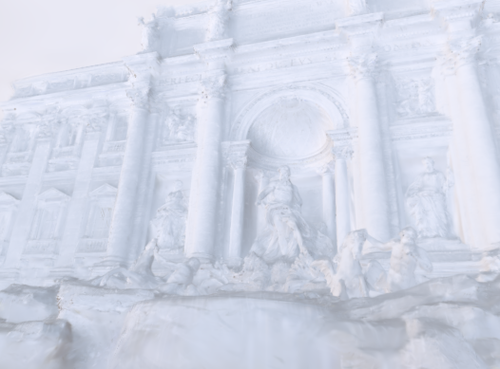} \\
        \inputimagerotate{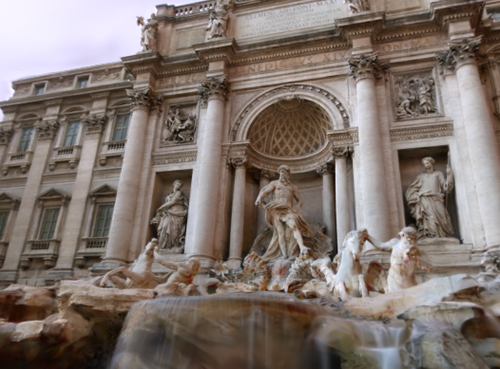} \&
        \inputimagerotate{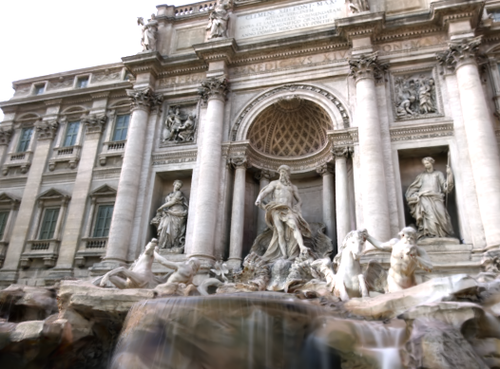} \&
        \inputimagerotate{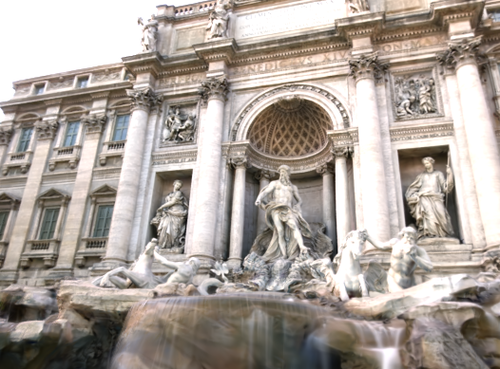} \\
        \inputimagerotate{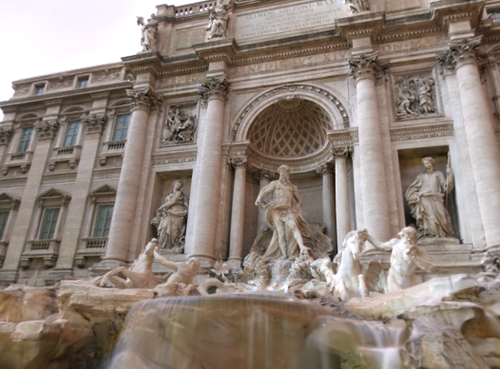} \&
        \inputimagerotate{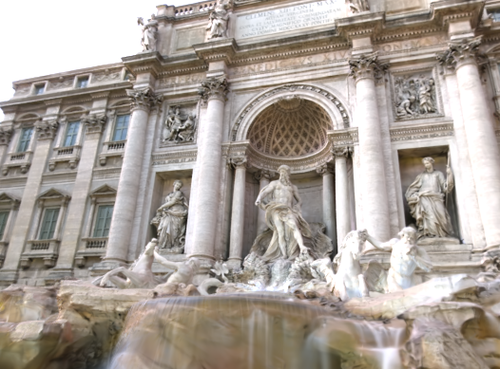} \&
        \inputimagerotate{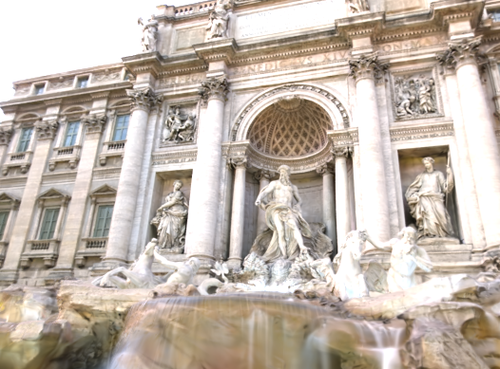} \\
        \inputimagerotate{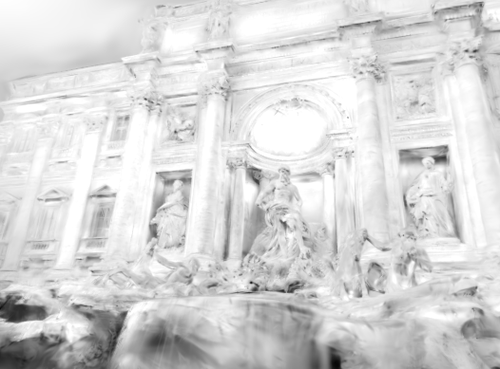} \&
        \inputimagerotate{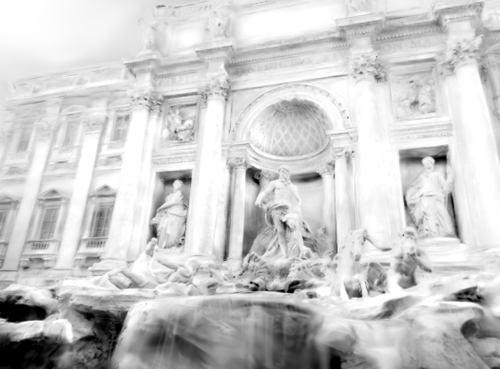} \&
        \inputimagerotate{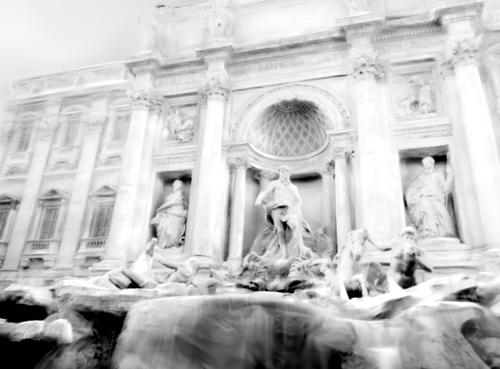} \\
    }; 
    
    \matrix[matrix of nodes, right=1em of scene1.east, column sep=0pt, row sep=0pt, ampersand replacement=\&, inner sep=0, outer sep=0] (scene2) {
        \inputimagerotate{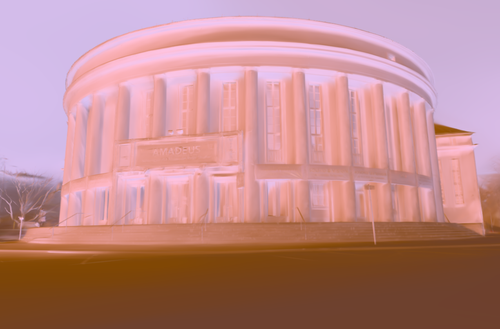} \&
        \inputimagerotate{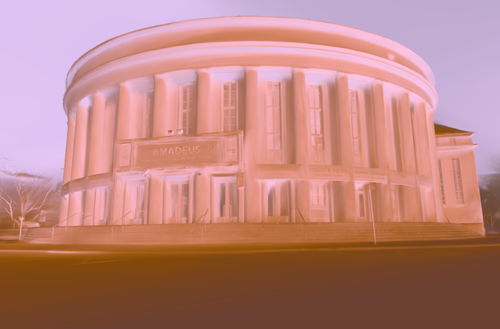} \&
        \inputimagerotate{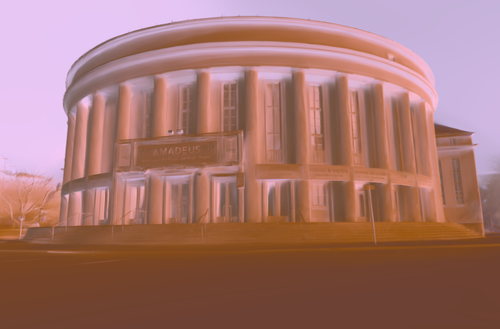} \\
        \inputimagerotate{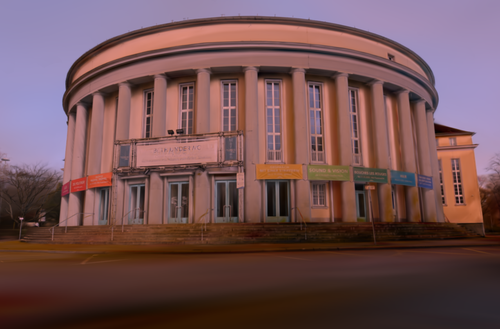} \&
        \inputimagerotate{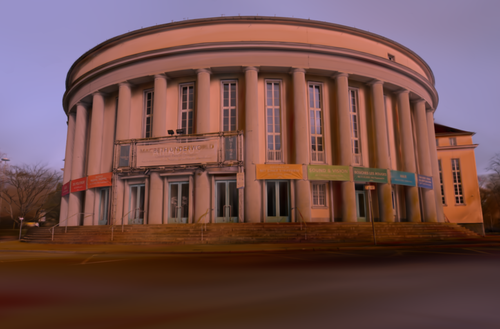} \&
        \inputimagerotate{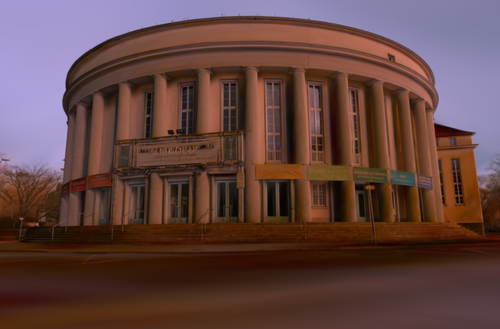} \\
        \inputimagerotate{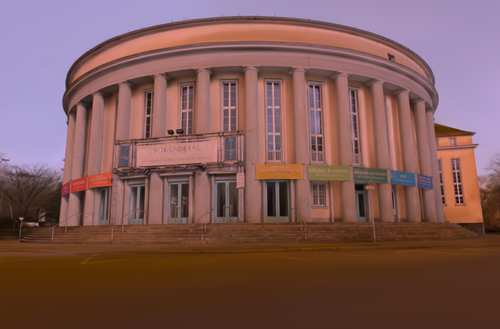} \&
        \inputimagerotate{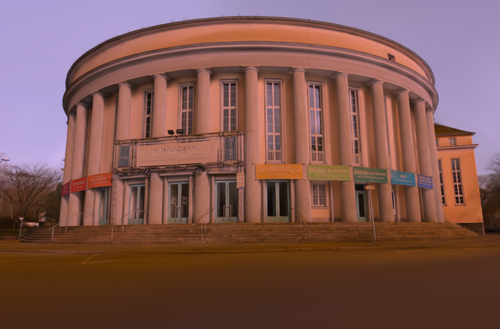} \&
        \inputimagerotate{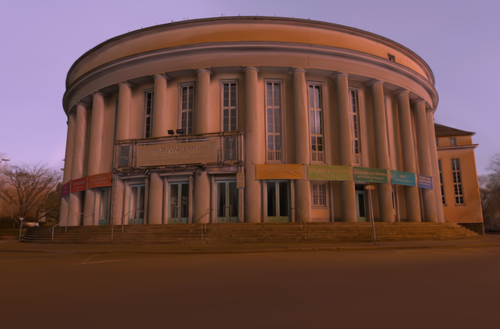} \\
        \inputimagerotate{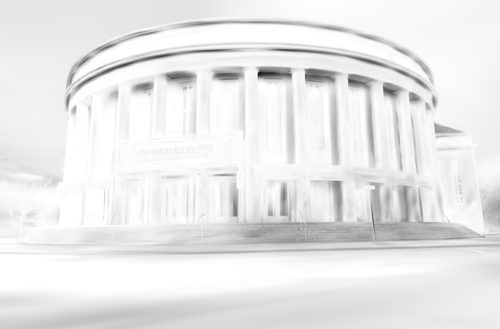} \&
        \inputimagerotate{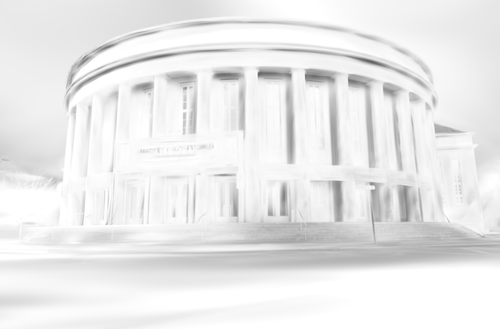} \&
        \inputimagerotate{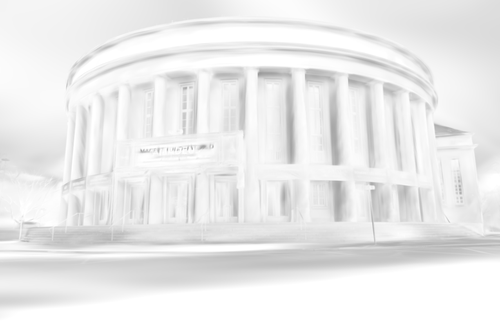} \\
    }; 
    \node[rotate=90, left=0.3em of scene1-1-1.west, anchor=base] {Illumination};
    \node[rotate=90, left=0.3em of scene1-2-1.west, anchor=base] {Shadowed};
    \node[rotate=90, left=0.3em of scene1-3-1.west, anchor=base] {Unshadowed};
    \node[rotate=90, left=0.3em of scene1-4-1.west, anchor=base] {Shadows};

    \node[anchor=north west] at (scene1-1-1.north west) {\includegraphics[width=32px]{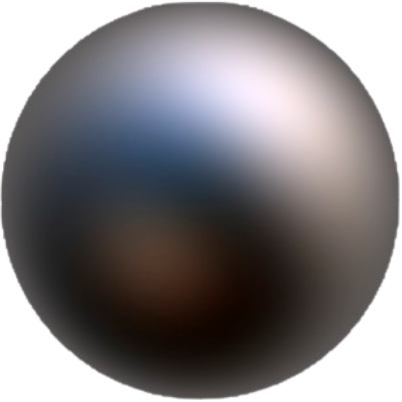}};
    \node[anchor=north west] at (scene2-1-1.north west) {\includegraphics[width=32px]{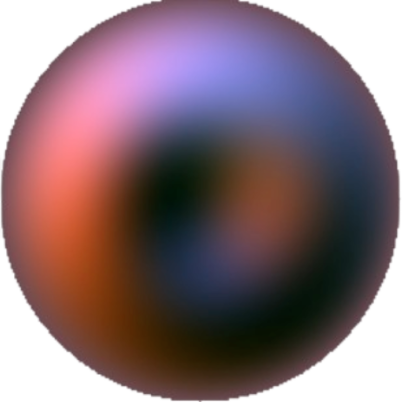}};
\end{tikzpicture}
}
}

\begin{figure*}[!t]
    \centering
    \rotatingfigure
    \caption{\textbf{Environment map rotation --} The top row shows the illumination entering the scene. The second and third rows display the shadowed and unshadowed renderings, respectively. The last row represents the approximate predicted shadows.
    Please zoom in for details.}
    \label{fig:rotate_light}
    \vspace{-1em}
\end{figure*}

\kk{As our backbone, 2DGS~\cite{huang20242dgs} incorportates priors to produce sharp edges and smooth surfaces, our model inherently performs better as expressed by SSIM. Please also see the zoom-ins in~\cref{fig:qual_albedo_norm_relight}. Those shape reconstruction qualities allow us to relight the scene with high fidelity.}
\change{We demonstrate that in~\cref{fig:rotate_light} where one can see that our method effectively relights landmarks under various lighting conditions. We finally visualize the rendered shadows produced thanks to our proposed physical constraints at training time. Since \ours{} does not predict shadows explicitly, we visualize them as grayscaled difference of output irradiances between the \textit{unshadowed} (\cref{eq:unshadowed-radiance}) and \textit{shadowed} (\cref{eq:shadowed-radiance}) to approximate shadow effects:
\begin{equation}
\max(\mathfrak{g}(\radiance_k \oslash \albedo_k - \shadowedradiance_k \oslash \albedo_k), 0),
\end{equation}
where $\oslash$ is an element-wise division and $\mathfrak{g}(\cdot)$ converts from the RGB space to the grayscale space. }

\change{We also display the illumination in~\cref{fig:rotate_light} to differentiate between shadows and dark illumination from the environment map.   
Additional detailed results on scene reconstruction, relightning and more comparisons to other works are included in~\supplementary{}}.

\subsection{Ablations}

We prioritize enhancing relighting capabilities over accurate appearance recreation during the optimization process, contrasting with recent Gaussian splatting methods that target novel view synthesis based on unconstrained photo collections~\cite{xu2024wildgs,dahmani2024swagsplattingwildimages,zhang2024gaussian}. Consequently, our ablation study primarily focuses on the degradation of relighting capabilities when removing any of the proposed components. We compare shadowed and unshadowed modeling and investigate the contributions of each loss term. We present the results in \cref{tab:results_osr_eval}.

Gaussians can optimize to shadowed surfaces and represent shadows as normals and albedo colors (effect known as albedo/illumination ambiguity). Therefore, gains from separating shadows from lightning are not visible in metrics computed on a limited data subset. We noticed that adding a shadowed version can help restore proper albedo and normal vectors of surfaces that during the training were distorted or had low brightness (see \cref{fig:shadowed_fix_normals}).

\subsection{Performance comparison}
\change{We compare \ours{}' efficiency with two NeRF baselines. Our method achieves plausible relightning results while being orders of magnitudes faster both in terms of training and inference as shown in \cref{tab:performance}.}
\subsection{Limitations}
We identify the following limitations of our approach. Notably, surface albedo and normals may attempt to simulate shadows in scenarios with hard and frequent shadows. This can pose challenges for shadow training, especially when shadows are visible in several training images, potentially hindering the accurate representation of surface normals. \change{Incorporating priors for environment light and shadowing could further enhance disentanglement and light transport modeling as presented in the concurrent NeuSky~\cite{gardner2023neusky}.} While we assume diffuse albedo, valid for most outdoor cases, shadows can appear unnaturally on reflective surfaces such as windows. Separate background optimization could enhance the synthesis of scenes with extensive sky areas. \change{Finally, our shadow modeling baked-in the spherical harmonics representations is non trivial to extend to dynamic applications, such as autonomous driving.}

\section{Conclusions}
\label{sec:conclusions}

    We present \ours{}---the method capable of decoupling environment lighting and albedo of objects from images \textit{in-the-wild}. To this end, we apply 2DGS~\cite{huang20242dgs} to reconstruct the object's surface accurately and then use our proposed training components that correctly disentangle light properties from the rendered colors. As we show in the experiments, our approach achieves better reconstruction results than the baselines. We also present that one of our contributions---modeling shadows via leveraging Spherical Harmonics properties---provides shadows of high fidelity that react appropriately to changing environment light. \ours{} is a novel approach in the direction of inverting the rendering process from images \textit{in-the-wild}, reconstructing high-quality scene properties without sacrificing the fidelity of the output.

\section*{Acknowledgments}
This research was supported by Microsoft Research through the EMEA PhD Scholarship Programme, National Science Centre, Poland (grant no. 2022/47/O/ST6/01407), the European Union’s Horizon 2020 research and innovation program (grant agreement no. 857533, Sano), the International Research Agendas program of the Foundation for Polish Science, co-financed by the European Regional Development Fund, and the Horizon Europe Programme (HORIZON-CL4-2022-HUMAN-02) under the project "ELIAS: European Lighthouse of AI for Sustainability" (GA no. 101120237).
We thank NVIDIA Corporation for providing access to GPUs via NVIDIA’s Academic Hardware Grants Program. 

\clearpage
{\small
\bibliographystyle{ieee_fullname}
\bibliography{egbib}
}
\clearpage
\appendix

\section{Dataset Processing}

\change{\textbf{Occluders.}
To exclude occluders from training images we use masks provided with OSR dataset~\cite{rudnev2022nerfosr}.}

\change{\textbf{Test set.}
We test our approach on 5 viewpoints for each scene, as it was originally proposed in \cite{rudnev2022nerfosr}. For testing, we use test masks provided by \cite{rudnev2022nerfosr} and we stricly follow their evaluation protocol. For SSIM, we report the average value over the segmentation mask, utilizing the scikit-image implementation with a window size of 5 and eroding the segmentation mask by the same window size to exclude the influence of pixels outside the mask on the metric value.}

\change{\textbf{Testing with ground truth environment map.}
The authors of \cite{gardner2023neusky} made an effort to recover steps for environment map preprocessing and alignment. The preprocessing step is available in their repository, accessible at: \url{https://github.com/JADGardner/neusky/blob/main/notebooks/nerfosr_envmaps.ipynb}. The detailed discussion on SOL-NeRF \cite{solnerf} approach to environment map alignment is included in the NeuSky main paper \cite{gardner2023neusky} and also confirmed with SOL-NeRF authors.}

\section{Implementation details}

The appearance embedding vector is set to a size of 24 dimensions. For predicting the environment map, we use MLP with 3 fully-connected layers of size 64. We trained all models for 40000 iterations, the first training stage is set to 20000 iterations. The learning rate for MLP and embedding is set to 0.002, which after first training stage is reduced to 0.0002. We train gaussian spherical harmonics with a learning rate of 0.002. We set the loss function weights as follows:
for $\loss{0-1}$ $\lossweight{1}=0.001$, for $\loss{+}$ $\lossweight{2}=0.05$, for $\loss{\moon{8}$\leftrightarrow$\moon{16}}$  $\lossweight{3} \in \{1.0, 10.0\}$, for $\loss{\moon{8}}$  $\lossweight{4}=10.0$. In the second training stage we set $\lossweight{\moon{8}}=0.001$.

\change{We adhere to the original Gaussian splatting densification and pruning protocols, with a densification interval of 500 iterations and an opacity reset interval of 3000 iterations. We apply regularizations to align Gaussians with surfaces, as originally described in \cite{huang20242dgs}. Additionally, we utilize the dual visibility concept proposed in \cite{huang20242dgs}. This ensures that the Gaussians are always correctly oriented towards the camera. Dual visibility effectively produces consistent world normals, with visible normals being consistent and non-visible ones contributing minimally to the rendering. Regularization of Spherical Harmonics $\mathbf{d}_{k}$ is dependent on gaussian normals. Since normals are rotated to always face the camera, to maintain alignment between each Gaussian's normal and its associated \( \mathbf{d}_k \), we also rotate \( \mathbf{d}_k \) accordingly.
}

\change{We run all experiments using a single NVIDIA A100 80GB or RTX 2080 Ti 128 GB.}

\section{Relighting - additional results}
Please reach for additional results to the attached videos (\url{https://drive.google.com/drive/folders/1AvKkg0MMWPsftFXMPoeuV3jCkmox3XxN?usp=sharing}).

\def\height{64px}
\def\spysize{24px}
\def\offset{36px}
\def\scale{0.8}
\def\scaleleft{0.8}
\def\distabove{0.5em}
\def\distleft{0.0em}
\def\clipleft{2cm}
\def\clipright{2cm}
\renewcommand{\inputimage}[1]{\includegraphics[height=\height,trim={8px 0 8px 0}, clip]{images/sep_images/qual_relight_albedo_norm/#1}}
\renewcommand{\inputimageshadows}[1]{\includegraphics[height=\height,trim={8px 0 8px 0}, clip]{images/sep_images/qual_gt_shadows/#1}}

\newcommand{\shadowsfigure}{
\centering
\resizebox{\linewidth}{!}{
    \begin{tikzpicture}[
        >=stealth',
        title/.style={
            anchor=base,
            align=center,
            scale=\scale,
        },
        spy using outlines={circle, red, magnification=3, connect spies, size=\spysize}
    ]
    
    \matrix[
        matrix of nodes, 
        anchor=west, 
        column sep=0pt, 
        row sep=0pt, 
        ampersand replacement=\&, 
        inner sep=0, 
        outer sep=0, 
    ] (shadows) {
        \inputimageshadows{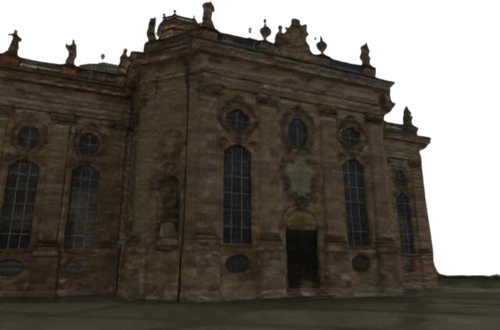} \&
        \inputimageshadows{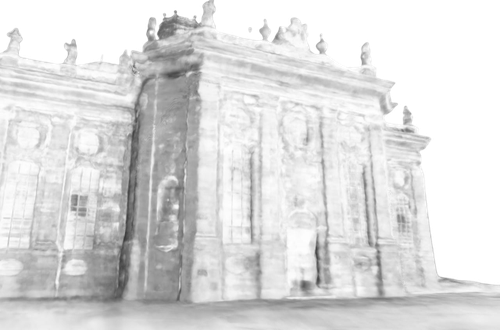} \\
        \inputimageshadows{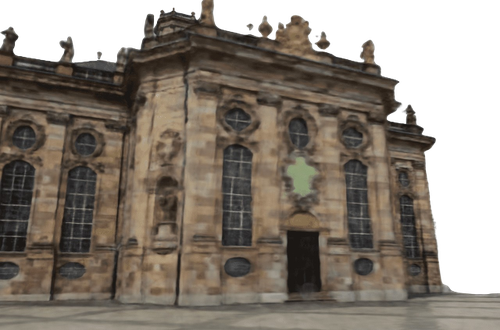} \&
        \inputimageshadows{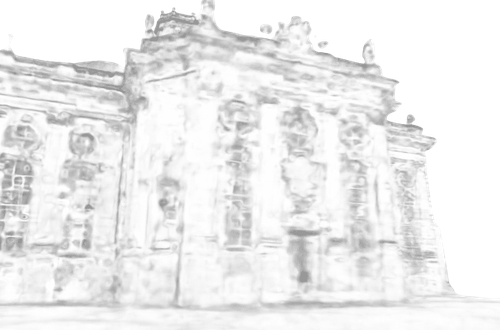} \\
        \inputimageshadows{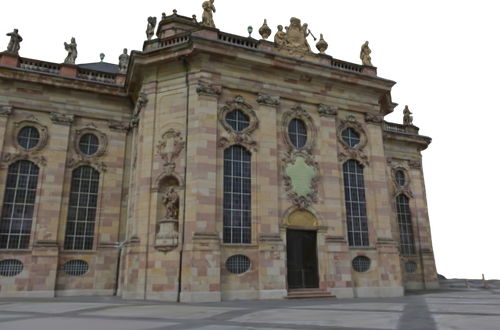} \&
        \inputimageshadows{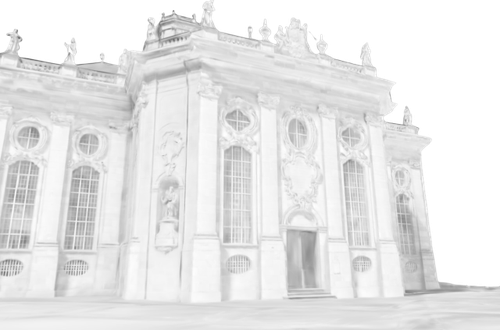} \\
        \inputimageshadows{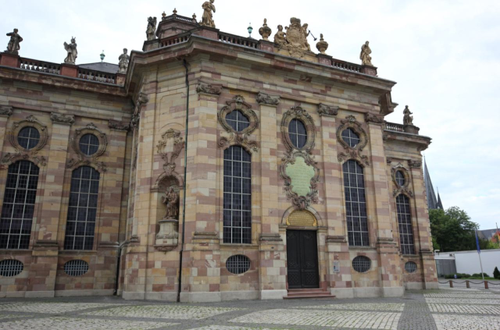} \\
    }; 
    
    \node[above=\distabove of shadows-1-1.north, title]{Relit Reconstruction};
    \node[above=\distabove of shadows-1-2.north, title]{Predicted Shadows};

    \node[left=\distleft of shadows-1-1.west, align=center,scale=\scaleleft]{\rotatebox{90}{NeRF-OSR~\cite{rudnev2022nerfosr}}};
    \node[left=\distleft of shadows-2-1.west, align=center,scale=\scaleleft]{\rotatebox{90}{SR-TensorRF~\cite{chang2024srtensorf}}};
    \node[left=\distleft of shadows-3-1.west, align=center,scale=\scaleleft]{\rotatebox{90}{\textbf{Ours}}};
    \node[left=\distleft of shadows-4-1.west, align=center,scale=\scaleleft]{\rotatebox{90}{Ground Truth}};

    \foreach \y in {1,...,3}{
        \foreach \x in {1,2}{
            \edef\temp{\noexpand\spy[anchor=center,magnification=2] on ([xshift=-1.6em,yshift=-0.7em] shadows-\y-\x.center) in node[below left=3.75px and 3.75px] at (shadows-\y-\x.north east);}
            \temp 
        }
    }
    \spy[anchor=center,magnification=2] on ([xshift=-1.6em,yshift=-0.7em] shadows-4-1.center) in node[below left=3.75px and 3.75px] at (shadows-4-1.north east);
\end{tikzpicture}
}
}

\begin{figure}[!t]
    \centering
    \shadowsfigure
    \caption{ \textbf{\change{Qualitative comparison of scene reconstruction for the selected photo session.}} Results for NeRF-OSR~\cite{rudnev2022nerfosr} and \ours{} were genrated using ground truth enviroment maps for selected photo session, while ST-TensoRF~\cite{chang2024srtensorf} used extracted timestamp. Results for NeRF-OSR and SR-TensorF reported originally in \cite{chang2024srtensorf}.} 
    \label{fig:qual_gt_shadows}
\end{figure}

\section{Qualitative comparison - additional results}
\change{In~\cref{fig:qual_gt_shadows} we show the qualitative comparison of our method, NeRF-OSR, and SR-TensoRF. We show the landmark relit with ground truth envoronment map for NeRF-OSR and \ours{}. SR-TensoRF reconstructs ground truth using only daytime (timestamp).}  

\change{In~\cref{fig:qualitative_appendix}, we show the qualitative comparison of our method, NeRF-OSR, and SR-TensoRF. We use the \textit{default synthetic} environment map provided by \cite{rudnev2022nerfosr}. This environment map was used for visualisation purposes in \cite{chang2024srtensorf}. We use it to ensure a fair comparison and consistency with results from concurrent works. We also present albedo and normals extracted from the reconstructed scene. Please note that our model produces much cleaner results. Compared to the baselines, it reconstructs sharp features in small elements of the buildings, which is also reflected in the quantitative results~\cref{tab:results_osr_eval}. \ours{} also gracefully smooths out the elements of scenes that are variable across the images, such as trees and clouds. On the other hand, NeRF-OSR and SR-TensoRF produce artifacts that negatively impact the output reconstructions.}

\change{In \cref{fig:qual_normal_albedo_appendix} we present additional comparison with concurrent works. We focus on normal and albedo quality.

\begin{figure*}[!b]
    \centering
    \includegraphics[width=\linewidth]{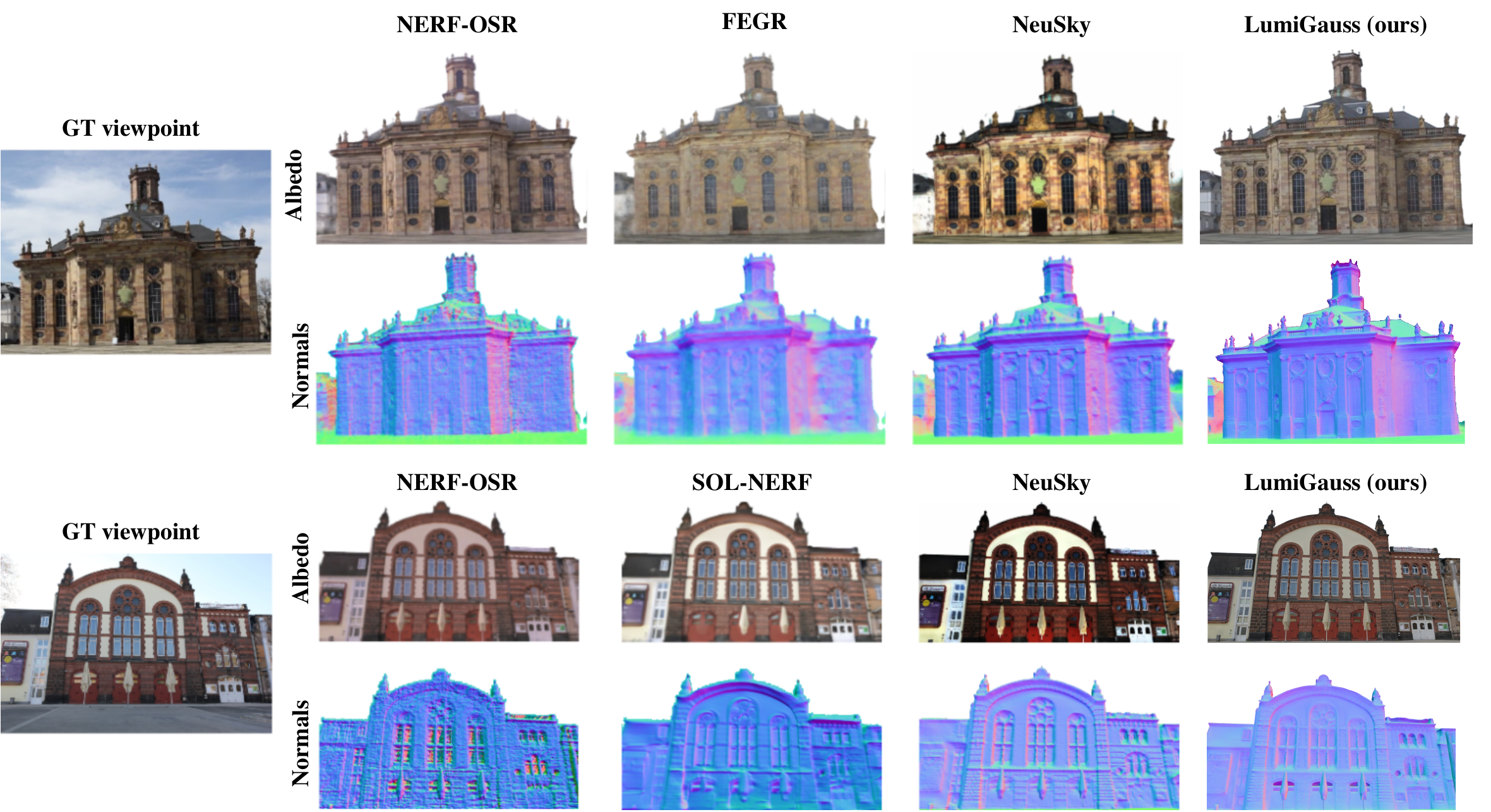}
    \caption{ \textbf{Qualitative comparison of predicted albedo and rendered normals.} Results for NERF-OSR, FEGR, SOL-NERF, NeuSky reported originally in \cite{gardner2023neusky}.} 
    \label{fig:qual_normal_albedo_appendix}
\end{figure*}
}

\begin{figure*}[!b]
    \centering
    \includegraphics[width=0.9\linewidth]{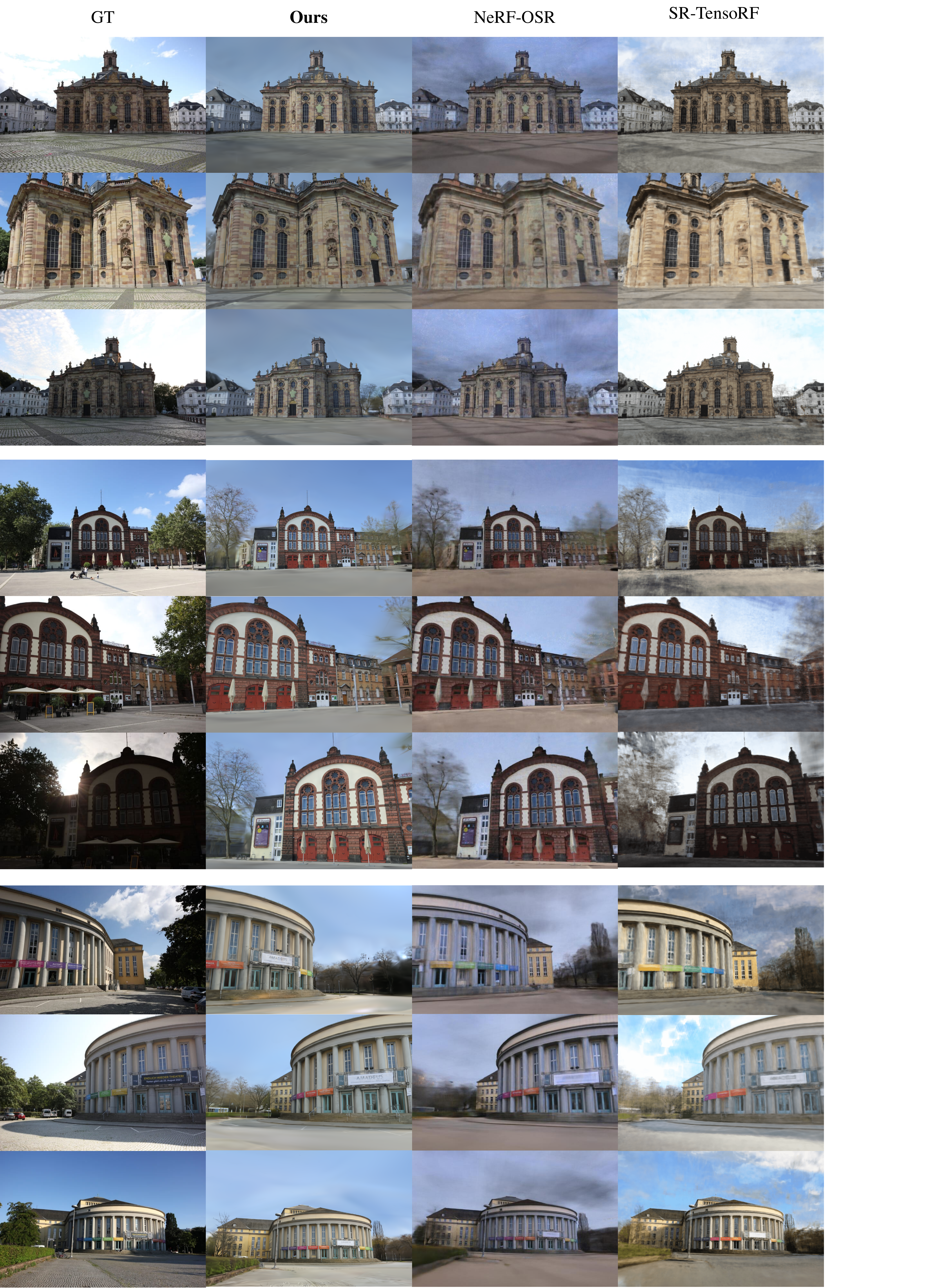}
    \caption{ \textbf{Qualitative comparison.} Additional novel viewpoints. Results for NeRF-OSR and SR-TensoRF originally reported in \cite{chang2024srtensorf}. Please note, in this comparison renders for NeRF-OSR and \ours{} \textbf{do not have to} reconstruct ground truth. \ours{} and NeRF-OSR employ the \textbf{default} environment map provided by NeRF-OSR \textbf{for clear visualisation purpose only}.} 
    \label{fig:qual_more_viewpoints_supp_img}
\end{figure*}

\change{In \cref{fig:qual_more_viewpoints_supp_img} we present additional results of novel view synthesis and comparison with concurrent works. Similarly to NeRF-OSR, we relight our scenes with the \textbf{default synthetic} map provided by NeRF-OSR for visualization purposes. This environment map does not correspond to GT images.}

\section{Ablation study - additional results }
\begin{figure*}[!b]
    \centering
    \includegraphics[width=\linewidth]{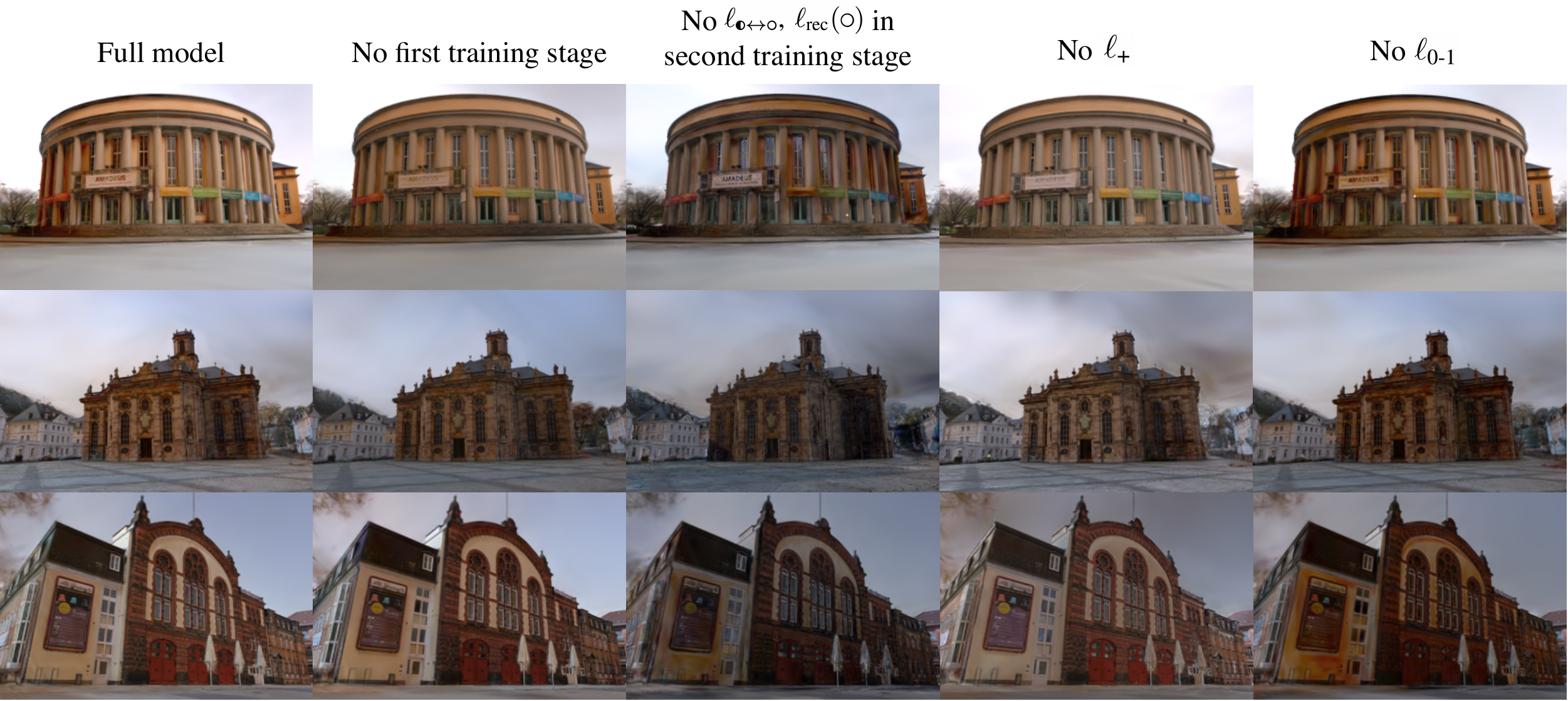}
    \caption{    \textbf{Ablation study for relightning with external environment map.} The full model results in the clearest render. The strongest quality drop is observed when components restricting $\transferfun_k$ are omitted.} 
    \label{fig:ablation_img}
\end{figure*}

In \cref{fig:ablation_img} we present renders from training without selected regularization terms.

\def\scale{1.0}
\def\distabove{0.5em}
\def\height{64px}
\def\spysize{24px}
\def\offset{36px}
\renewcommand{\inputimage}[1]{\includegraphics[height=\height]{images/sep_images/qual/#1}}

\newcommand{\qualitative}{
\centering
\resizebox{\linewidth}{!}{
    \begin{tikzpicture}[
        >=stealth',
        title/.style={
            anchor=base,
            align=center,
            scale=\scale,
        },
        spy using outlines={circle, red, magnification=3, connect spies, size=\spysize}
    ]
    \matrix[matrix of nodes, column sep=0pt, row sep=0pt, ampersand replacement=\&, inner sep=0, outer sep=0, font=\Huge] (qualitative) {
        \inputimage{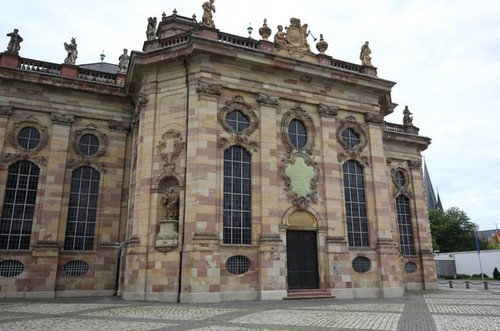} \&
        \inputimage{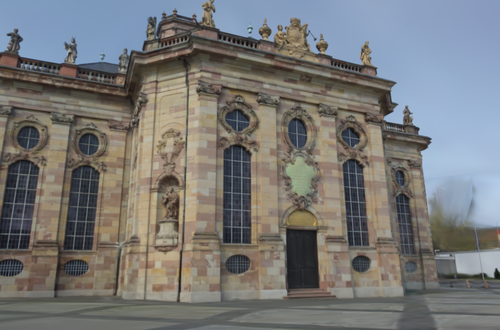} \&
        \inputimage{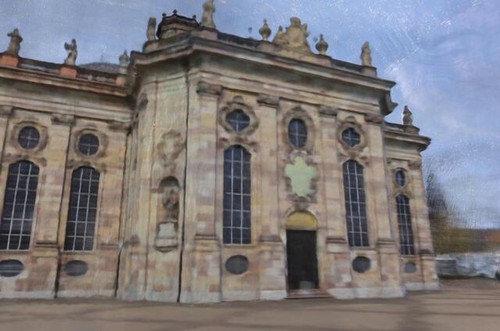} \&
        \inputimage{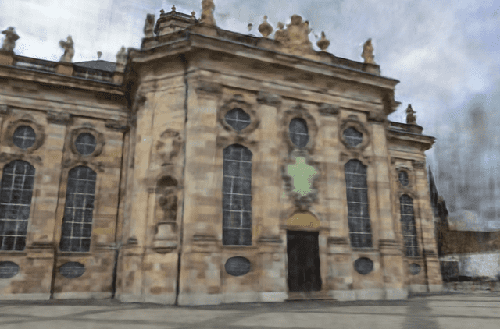} \\
        \inputimage{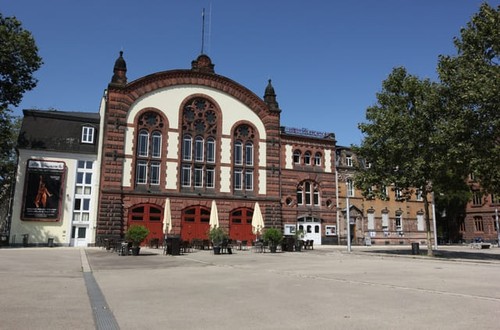} \&
        \inputimage{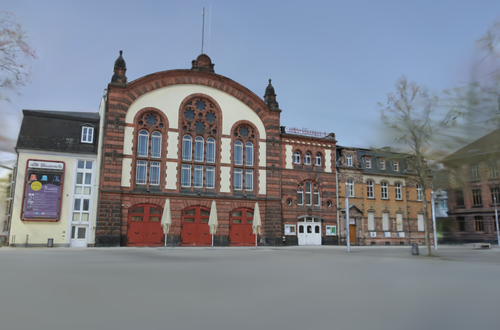} \&
        \inputimage{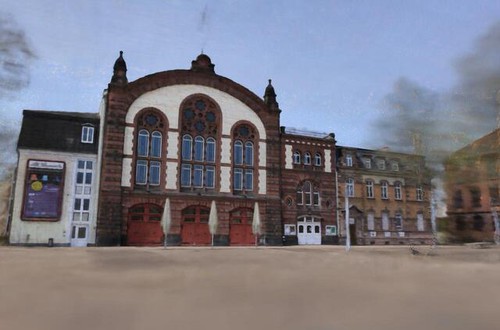} \&
        \inputimage{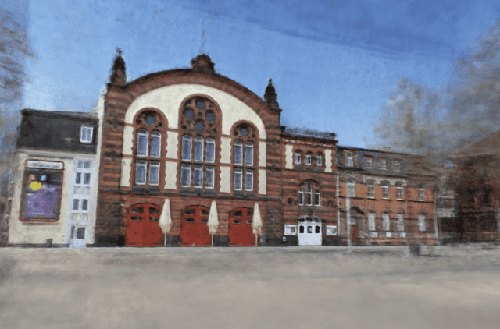} \\
        \inputimage{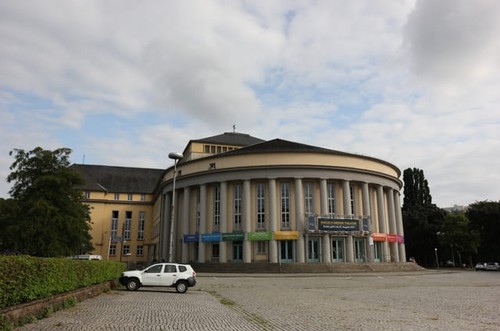} \&
        \inputimage{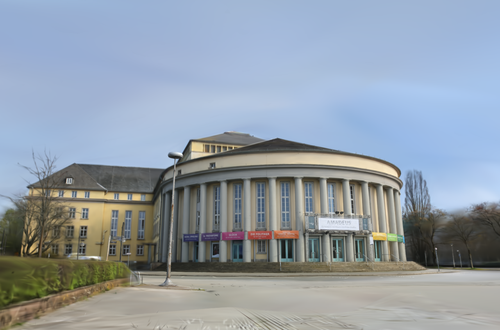} \&
        \inputimage{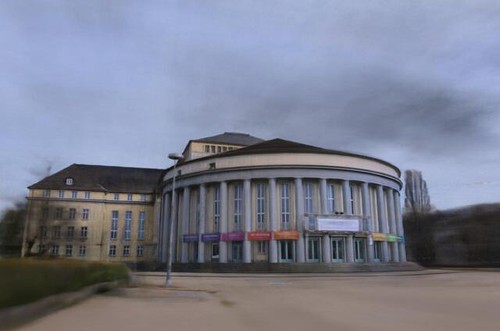} \&
        \inputimage{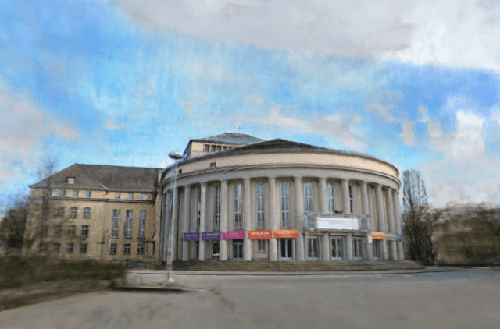} \\
    }; 
    
    \matrix[
        right=16px of qualitative.east, 
        matrix of nodes, 
        anchor=west, 
        column sep=0pt, 
        row sep=0pt, 
        ampersand replacement=\&, 
        inner sep=0, 
        outer sep=0, 
    ] (ours) {
        \inputimage{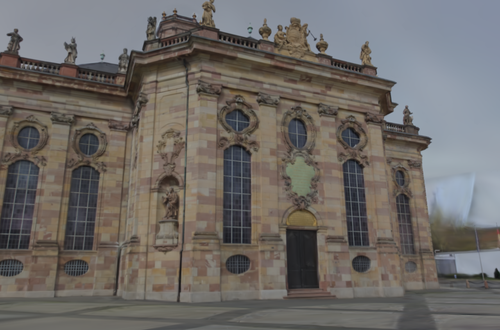} \&
        \inputimage{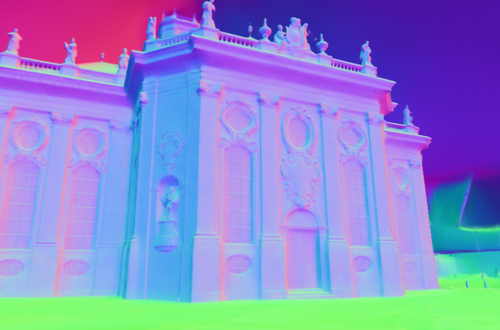} \\
        \inputimage{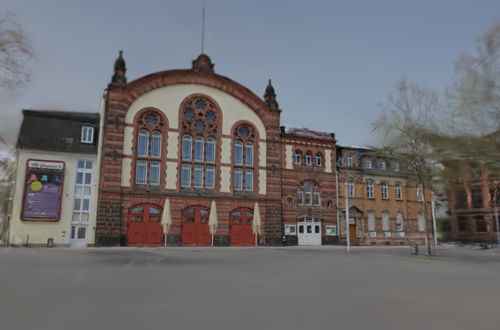} \&
        \inputimage{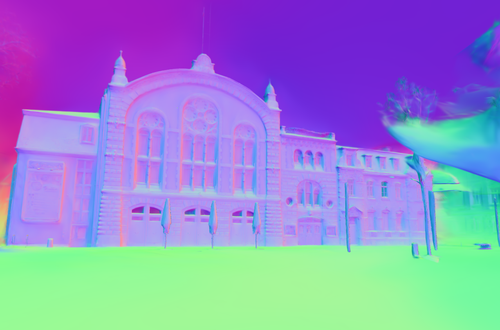} \\
        \inputimage{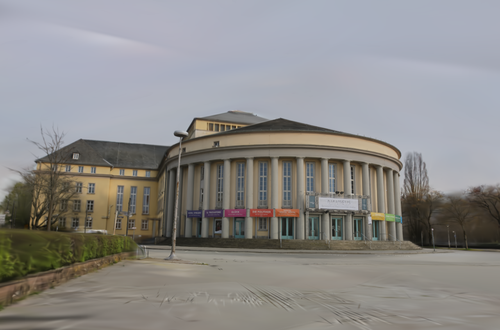} \&
        \inputimage{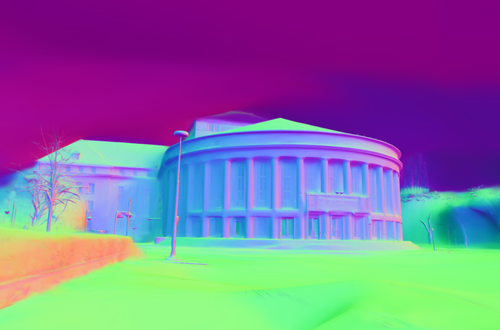} \\
    }; 

    \node[above=\distabove of qualitative-1-1.north, title] {Ground Truth};
    \node[above=\distabove of qualitative-1-2.north, title] {\textbf{Ours}};
    \node[above=\distabove of qualitative-1-3.north, title] {NeRF-OSR~\cite{rudnev2022nerfosr}};
    \node[above=\distabove of qualitative-1-4.north, title] {SRTensoRF~\cite{chang2024srtensorf}};
    
    \node[above=\distabove of ours-1-1.north, title] {\textbf{Ours} Albedo};
    \node[above=\distabove of ours-1-2.north, title] {\textbf{Ours} Normals};

    \spy[anchor=center] on ([xshift=1em] qualitative-1-1.center) in node[below left=3.75px and 3.75px] at (qualitative-1-1.north east);
    \spy[anchor=center] on ([xshift=1em] qualitative-1-2.center) in node[below left=3.75px and 3.75px] at (qualitative-1-2.north east);
    \spy[anchor=center] on ([xshift=1em] qualitative-1-3.center) in node[below left=3.75px and 3.75px] at (qualitative-1-3.north east);
    \spy[anchor=center] on ([xshift=1em] qualitative-1-4.center) in node[below left=3.75px and 3.75px] at (qualitative-1-4.north east);
    
    \spy[anchor=center] on ([xshift=-1em] qualitative-2-1.center) in node[below left=3.75px and 3.75px] at (qualitative-2-1.north east);
    \spy[anchor=center] on ([xshift=-1em] qualitative-2-2.center) in node[below left=3.75px and 3.75px] at (qualitative-2-2.north east);
    \spy[anchor=center] on ([xshift=-1em] qualitative-2-3.center) in node[below left=3.75px and 3.75px] at (qualitative-2-3.north east);
    \spy[anchor=center] on ([xshift=-1em] qualitative-2-4.center) in node[below left=3.75px and 3.75px] at (qualitative-2-4.north east);
    
    \spy[anchor=center] on ([xshift=-1em] qualitative-3-1.center) in node[below left=3.75px and 3.75px] at (qualitative-3-1.north east);
    \spy[anchor=center] on ([xshift=-1em] qualitative-3-2.center) in node[below left=3.75px and 3.75px] at (qualitative-3-2.north east);
    \spy[anchor=center] on ([xshift=-1em] qualitative-3-3.center) in node[below left=3.75px and 3.75px] at (qualitative-3-3.north east);
    \spy[anchor=center] on ([xshift=-1em] qualitative-3-4.center) in node[below left=3.75px and 3.75px] at (qualitative-3-4.north east);
\end{tikzpicture}
}
}

\begin{figure*}[!t]
    \centering
    \qualitative
    \caption{%
            \textbf{Qualitative results --} Showcase of novel view synthesis using shadowed radiance transfer. We present albedo and normals produced by our method. Our method generates much \change{sharper} renderings. Please see zoom-ins to see details on the quality difference, such as surface smoothness and edge sharpness of small building elements. We use visual results for SR-TensoRF and NeRF-OSR presented originally in~\cite{chang2024srtensorf}. 
            \change{Please note that, in this comparison the environment map used to create renders for NeRF-OSR and \ours{} \textbf{does not match} the illumination in ground truth. \ours{} and NeRF-OSR employ the \textbf{default} environment map provided by NeRF-OSR \textbf{for clear visualisation purpose only}. SR-TensoRF do not rely on any environment map, instead it utilizes daytime information}.             
        }
    
    \label{fig:qualitative_appendix}
\end{figure*}

\end{document}